\documentclass[fleqn,10pt]{wlscirep}
\usepackage[utf8]{inputenc}
\usepackage[T1]{fontenc}

\usepackage{import}
\usepackage{amsmath,graphicx}
\usepackage{amssymb}
\usepackage{amsfonts}
\usepackage{xcolor}
\usepackage{tabularx,booktabs}
\usepackage{makecell}
\usepackage{colortbl}
\usepackage{multirow}
\usepackage{multicol}
\usepackage{array}
\usepackage{adjustbox}
\usepackage{caption}
\usepackage{subcaption}
\usepackage{float}
\usepackage{verbatim}
\usepackage{url}
\usepackage{hyperref}
\hypersetup{
    colorlinks,
    linkcolor={red!50!black},
    citecolor={blue!50!black},
    urlcolor={blue}
}

\definecolor{darkblue}{HTML}{0173B2}
\definecolor{lightblue}{HTML}{56B4E9}

\usepackage{pgf}

\usepackage{tikz}
\usetikzlibrary{positioning}
\usetikzlibrary{decorations.pathreplacing}
\usetikzlibrary{calc}

\newcommand{\includepgf}[1]{%
  \IfFileExists{#1.pdf}{\includegraphics{#1.pdf}}{\input{#1.pgf}}%
}

\usepackage{layouts}
\usepackage{pdfpages}

\title{A benchmark for video-based laparoscopic skill analysis and assessment}

\author[1,2]{Isabel Funke}
\author[1,2]{Sebastian Bodenstedt}
\author[3,4]{Felix von Bechtolsheim}
\author[3,4]{Florian Oehme}
\author[3]{Michael Maruschke}
\author[3]{Stefanie Herrlich}
\author[3]{Jürgen Weitz}
\author[3,4]{Marius Distler}
\author[5]{Sören Torge Mees}
\author[1,2]{Stefanie Speidel}

\affil[1]{Department of Translational Surgical Oncology, National Center for Tumor Diseases (NCT), NCT/UCC Dresden, a partnership between DKFZ, Faculty of Medicine and University Hospital Carl Gustav Carus, TUD Dresden University of Technology, and Helmholtz-Zentrum Dresden-Rossendorf (HZDR), Germany}
\affil[2]{The Centre for Tactile Internet with Human-in-the-Loop (CeTI), TUD Dresden University of Technology, Dresden, Germany}
\affil[3]{Department for Visceral, Thoracic and Vascular Surgery, Medical Faculty and University Hospital Carl Gustav Carus, TUD Dresden University of Technology, Dresden, Germany}
\affil[4]{Surgical Skills Lab Dresden, Medical Faculty and University Hospital Carl Gustav Carus, TUD Dresden University of Technology, Dresden, Germany}
\affil[5]{Department of General and Visceral Surgery, Municipal Hospital Dresden-Friedrichstadt, Dresden, Germany}

\begin{abstract}

Laparoscopic surgery is a complex surgical technique that requires extensive training. Recent advances in deep learning have shown promise in supporting this training by enabling automatic video-based assessment of surgical skills. 
However, the development and evaluation of deep learning models is currently hindered by the limited size of available annotated datasets.

To address this gap, we introduce the Laparoscopic Skill Analysis and Assessment (LASANA) dataset, comprising 1270 stereo video recordings of four basic laparoscopic training tasks. Each recording is annotated with a structured skill rating, aggregated from three independent raters, as well as binary labels indicating the presence or absence of task-specific errors.
The majority of recordings originate from a laparoscopic training course, thereby reflecting a natural variation in the skill of participants. 

To facilitate benchmarking of both existing and novel approaches for video-based skill assessment and error recognition, we provide predefined data splits for each task. Furthermore, we present baseline results from a deep learning model as a reference point for future comparisons.

\end{abstract}

\begin{document}

\flushbottom
\maketitle

\section*{Background \& Summary}

Laparoscopic surgery offers benefits for patients, including reduced blood loss, less postoperative pain, and faster recovery. The surgical technique refers to procedures where the surgeon operates on abdominal or pelvic organs using long, slender instruments inserted through small incisions in the patient's body. For real-time visual information about the procedure, an endoscopic camera, introduced through a separate incision, provides a video feed of the surgical field. % , which is displayed on a monitor. 
% To create sufficient working space and enhance visibility, the patient’s abdominal cavity is insufflated with gas.

Despite advantages for patients, laparoscopic surgery is demanding for surgeons\cite{supe2010ergonomics}: While manipulating the instruments, surgeons must look at a two-dimensional video display to observe the procedure, resulting in a decoupling of the visual and motor axes. Compared to open surgery, surgeons also lose three-dimensional peripheral vision and have impaired tactile feedback. Moreover, the instruments are constrained by fixed entry points % through the abdominal wall 
so that their movement is limited. % to rotations around these access sites. 
This introduces the fulcrum effect \cite{gallagher1998ergonomic}, where the tip of the instrument moves in the opposite direction of the surgeon’s hand.

To acquire laparoscopic skills, novice surgeons typically begin training in a simulated dry-lab environment: A laparoscopic training box replicates the patient’s abdomen, allowing trainees to practice with inexpensive objects or materials designed to mimic real surgical scenarios. Dedicated training curricula comprise exercises that target core laparoscopic skills, such as object manipulation, cutting, and suturing. One example is the \emph{McGill Inanimate System for Training and Evaluation of Laparoscopic Skills (MISTELS)}\cite{fried2004proving}, which forms a key component of the \emph{Fundamentals of Laparoscopic Surgery (FLS)}\footnote{\url{https://www.flsprogram.org}} program.

Effective skills training relies on continuous assessment to provide feedback, monitor progress, sustain motivation, and determine readiness for transitioning to the operating room. Traditionally, feedback and evaluation are offered informally by experienced surgeons, often during dedicated laparoscopic training courses. However, this approach is limited by the availability of expert mentors, and informal human judgment is prone to subjectivity and noise \cite{kahnemann2021noise}.

To make manual surgical skill assessment more objective and reliable, structured evaluation frameworks have been proposed, such as checklists and global rating scales \cite{martin1997objective}. One example is the \emph{Global Operative Assessment of Laparoscopic Skills (GOALS)} \cite{vassiliou2005global} tool, which evaluates laparoscopic performance across five dimensions: depth perception, bimanual dexterity, efficiency, tissue handling, and autonomy. Each dimension is rated on a five-point Likert scale with defined anchors at points 1, 3, and 5, and the overall GOALS score is computed as the sum across all five dimensions.

Other assessment methods adopt a more quantitative approach. For example, MISTELS suggests to evaluate laparoscopic tasks primarily based on completion time, with penalties applied for specific errors or lack of precision \cite{derossis1998development}. Other studies have explored motion-based analysis, where descriptive parameters such as instrument path length, movement count, and idle time are extracted from instrument trajectories to quantify skill \cite{mason2013motion}. However, this approach requires dedicated tracking hardware unless instrument motion can be derived accurately from available surgical video \cite{law2017surgeon}. In robot-assisted surgery, by contrast, robot kinematics and system events data are inherently available and have been used to compute automated performance metrics \cite{guerin2022review}.

Clearly defined performance metrics for surgical skill assessment are promising because they are objective, automatic, and capable of delivering timely, consistent, and cost-effective feedback. Nevertheless, such metrics may appear rigid and may capture only limited aspects of performance, failing to fully reflect the nuanced judgments of expert raters. For this reason, recent research has investigated machine learning-based methods trained to infer human ratings, such as the total GOALS score, from recorded surgical task data. Early studies focused on classifying discrete skill levels (novice, intermediate, expert) or estimating overall skill scores using robot kinematic data \cite{zia2018automated,ismail2018evaluating,wang2018deep,anh2020towards}.

More recently, video has emerged as a rich and universally applicable modality for skill assessment. Video recordings provide comprehensive information even for retrospective evaluation by human raters \cite{mcqueen2019video}, and unlike kinematic data, they are readily available in both robotic and conventional laparoscopic procedures. Several studies have applied deep learning-based computer vision techniques to automatically assess skill directly from surgical videos, demonstrating promising performance \cite{funke2019video,wang2020towards,liu2021towards,anastasiou2023keep}.

Despite the progress in automatic laparoscopic skill assessment, however, currently available datasets for training and evaluating the recently proposed deep learning models remain limited. Most existing datasets focus on robot-assisted or open surgery, and/or are too small to enable comprehensive evaluation based on representative and diverse training and test data. Table~\ref{tab:datasets} summarizes the characteristics of existing datasets. 

\begin{table}[tb]
\newcolumntype{g}{>{\columncolor{gray!10}}l}
\renewcommand{\arraystretch}{1.75}
\caption{Overview of surgical training video datasets.}
\label{tab:datasets}
\centering
\begin{tabular}{gllll}
\rowcolor{gray!10}
Dataset & JIGSAWS \cite{gao2014jhu,ahmidi2017dataset} & ROSMA \cite{rivas2020training,rivas2023surgical} & AIxSuture \cite{peters2023aixsuture,hoffmann2024aixsuture} & \makecell[tl]{LASANA \cite{funke2026lasana} \\\emph{(this paper)}} \\
Training setting & \makecell[tl]{da Vinci \\Surgical System} & \makecell[tl]{da Vinci \\Research Kit} & \makecell[tl]{Simulated \\open surgery} & \makecell[tl]{Laparoscopic \\training box} \\
Tasks & \makecell[tl]{Suturing \\Knot tying \\Needle passing} & \makecell[tl]{Post and sleeve \\Pea on a peg \\Wire chaser} & Suturing & \makecell[tl]{Peg transfer\\ Circle cutting\\ Balloon resection\\ Suture \& knot} \\
Recorded data & \makecell[tl]{Stereo video \\Robot kinematics} & \makecell[tl]{Video \\Robot kinematics} & Video & Stereo video \\
Number of participants & 8 & 12 & 157 & 70 \\
Number of videos (total) & 103 & 206 & 314 & 1270 \\
Number of videos per task & 39; 36; 28 & 65; 71; 70 & 314 & 329; 311; 316; 314 \\
Mean video duration per task & \makecell[tl]{1\,min 53\,s; 57\,s; \\ 1\,min 48\,s} & \makecell[tl]{2\,min 10\,s; 2\,min 15\,s; \\ 36\,s} & 5 min & \makecell[tl]{2\,min 32\,s; 3\,min 32\,s; \\ 3\,min 55\,s; 4\,min 30\,s} \\
Manual annotations & \makecell[tl]{Experience level \\Skill rating \\Surgical gestures} & Task-specific errors & \makecell[tl]{Skill rating\\ Number of sutures} & \makecell[tl]{Skill rating\\ Task-specific errors} \\
Number of ratings per video & 1 & 1 & 3 & 3 \\            
\end{tabular}
\end{table}

To address this problem, we introduce the \emph{\textbf{La}paroscopic \textbf{S}kill \textbf{An}alysis and \textbf{A}ssessment (LASANA)} video dataset, a benchmark intended for the development and evaluation of methods for automatic video-based surgical skill analysis. LASANA comprises a total of 1270 stereo video recordings of four basic laparoscopic training tasks (peg transfer, circle cutting, balloon resection, and suture \& knot), performed by 70 participants. Each recording is annotated with a GOALS-inspired structured skill rating, aggregated across three independent raters to improve reliability. The average pairwise inter-rater agreement, quantified by Lin’s Concordance Correlation Coefficient~$\rho_c$ \cite{lawrence1989concordance}, exceeds 0.65 for all tasks except for circle cutting ($\rho_c = 0.49$).

To represent a broad spectrum of surgical skill, 58~participants were recorded multiple times throughout a laparoscopic training course so that the benchmark captures natural skill progression. In contrast, the AIxSuture dataset \cite{peters2023aixsuture} includes only two recordings per participant (one before and one after suturing training) whereas datasets like JIGSAWS \cite{gao2014jhu} and ROSMA \cite{rivas2020training} include several recordings per participant, but apparently without ongoing formal training.

In addition to skill ratings, LASANA provides annotations for task-specific errors such as dropping an object during the peg transfer task or puncturing the inner balloon during the balloon resection task. These annotations enable the development of automatic video-based error recognition algorithms to assess task performance from a complementary perspective.

To facilitate reproducible benchmarking, we define data splits for each task, dividing recordings into training, validation, and test sets at the level of participants (``leave several users out'' \cite{ahmidi2017dataset}). To establish a foundation for future comparisons, we report baseline results for overall skill score estimation and error recognition. For error recognition, this includes one representative error per task.
We anticipate that LASANA will serve as a valuable resource, enabling systematic benchmarking and fostering the development of advanced deep learning methods for automatic video-based skill analysis in laparoscopic surgery.

\section*{Methods}

\subsection*{Video recording}

\begin{figure}[htb]
    \centering
    \begin{tikzpicture} 
        \node[above right, inner sep=0] (image) at (0,0) {
            \includegraphics[width=0.4\linewidth]{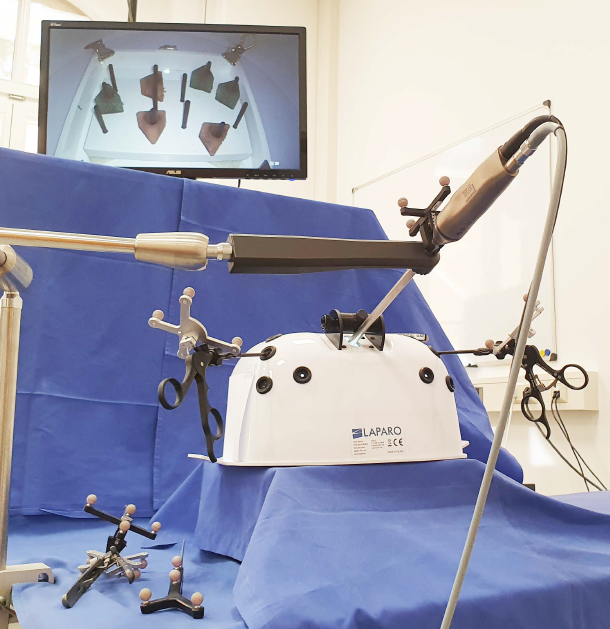}
        }; 

        % Create scope with normalized axes
        \begin{scope}[
            x={($0.1*(image.south east)$)},
            y={($0.1*(image.north west)$)}
        ]         
            % Grid
            % \draw[lightgray,step=1] (image.south west) grid (image.north east);

            % Axes' labels
            % \foreach \x in {0,1,...,10} { \node [below] at (\x,0) {\x}; }
            % \foreach \y in {0,1,...,10} { \node [left] at (0,\y) {\y}; }

            \node[circle,fill=white] at (5,2.7){\textsf{A}};
            \node[circle,fill=white] at (8,8){\textsf{B}};
            \node[circle,fill=white] at (5,9){\textsf{C}};
            \node[circle,fill=white] at (3,3){\textsf{D}};
            \node[circle,fill=white] at (9.4,3.4){\textsf{D}};
        \end{scope}
        
    \end{tikzpicture}
    \caption{Experimental setup for recording laparoscopic training tasks. The setup includes (A) a laparoscopic training box containing task-specific objects and materials, (B) a stereo endoscope, (C) a 2D monitor displaying the live video feed from the endoscope's left camera, and (D) standard laparoscopic instruments.}
    \label{fig:recording_setup}
\end{figure}
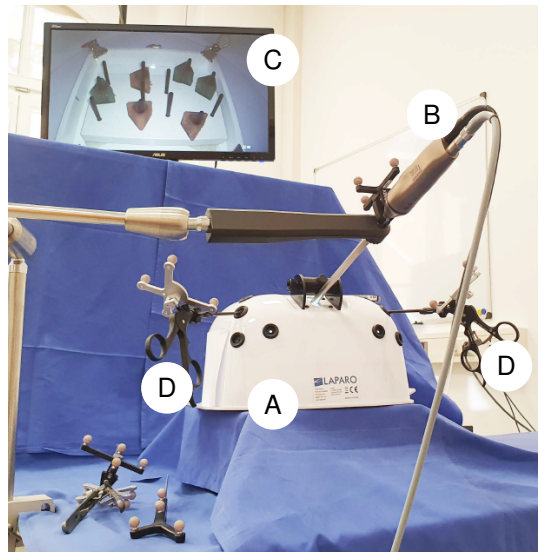

The LASANA dataset comprises recordings of four distinct laparoscopic training tasks, performed in a Laparo Aspire training box using laparoscopic instruments by Karl Storz. Synchronized stereo videos of the simulated surgical scene were captured with a Karl Storz TIPCAM 1 S 3D LAP 30° endoscope. Here, the video stream from the left camera was displayed on a monitor to provide visual feedback when performing a laparoscopic task. The experimental setup is depicted in Fig.~\ref{fig:recording_setup}.

\begin{figure}[tb]
  \centering

  \subcaptionbox{Peg transfer \label{fig:peg_transfer}}[0.35\linewidth]{%
    \includegraphics[width=\linewidth]{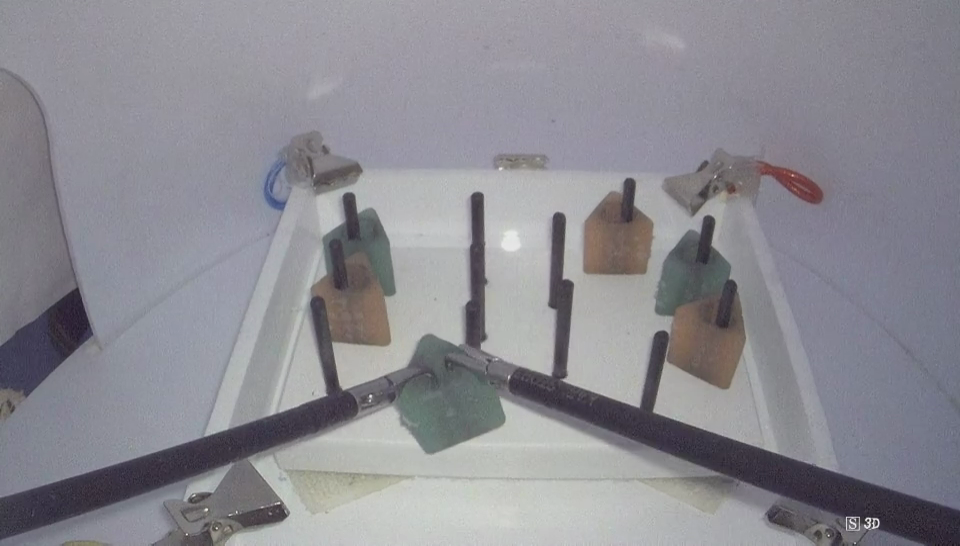}
  }
  \hspace{1em}
  \subcaptionbox{Circle cutting \label{fig:circle_cutting}}[0.35\linewidth]{%
    \includegraphics[width=\linewidth]{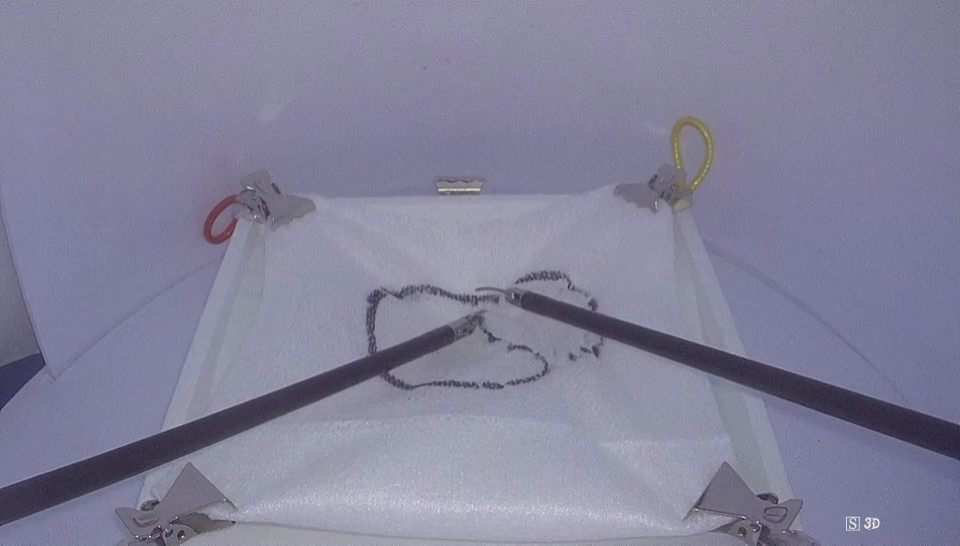}
  }

  \vspace{1em}

  \subcaptionbox{Balloon resection \label{fig:balloon_resection}}[0.35\linewidth]{%
    \includegraphics[width=\linewidth]{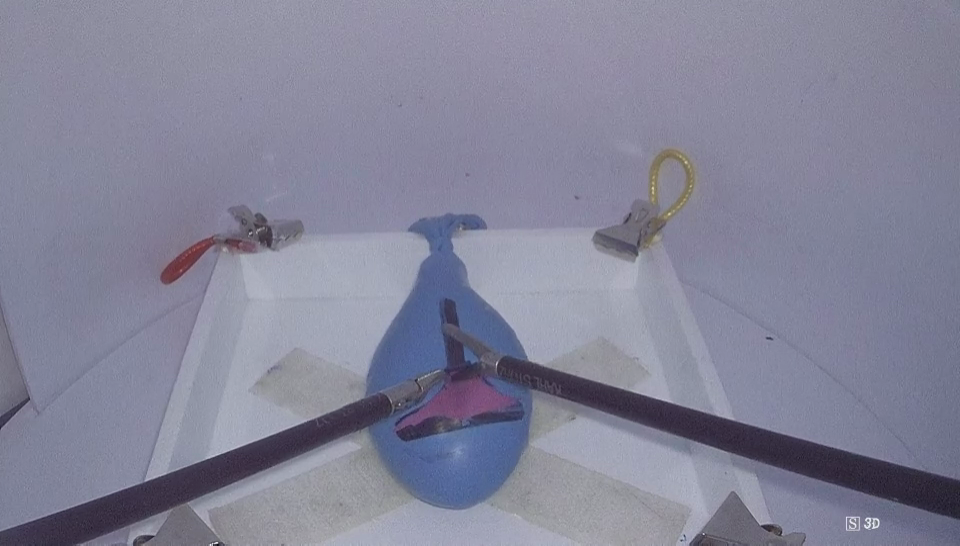}
  }
  \hspace{1em}
  \subcaptionbox{Suture \& knot \label{fig:suture_and_knot}}[0.35\linewidth]{%
    \includegraphics[width=\linewidth]{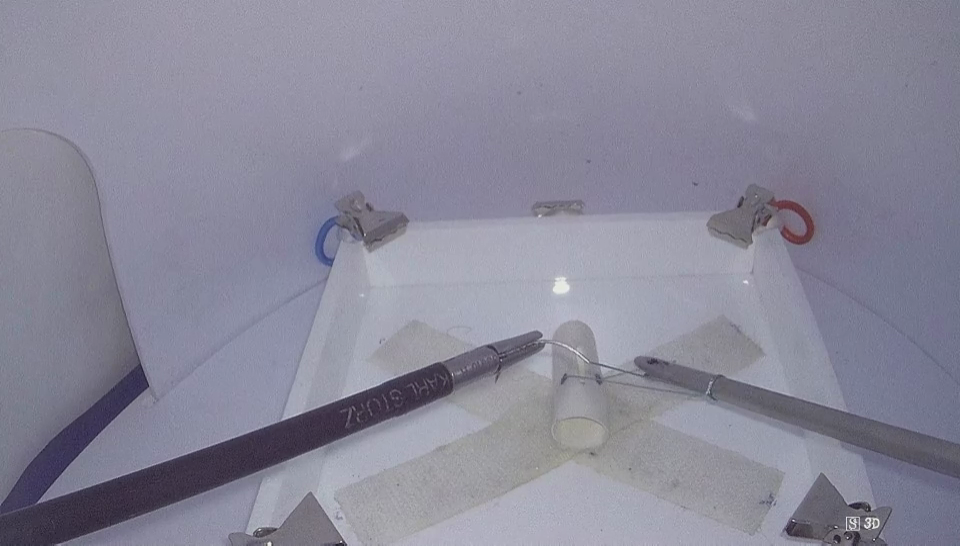}
  }

  \caption{The recorded laparoscopic training tasks.}
  \label{fig:training_tasks}
\end{figure}

Each recording consists of a left and a right video stream, both with a spatial resolution of $960 \times 540$ pixels and a frame rate of 20~frames per second \emph{(fps)}. The videos are encoded using the H.264 codec and stored in the Matroska (.mkv) container format. 
The following laparoscopic tasks were recorded:
\begin{itemize}[parsep=0pt]
    \item Peg transfer: Transfer six triangular objects from the left to the right side of a pegboard, then transfer them back.
    \item Circle cutting: Accurately cut along a pre-marked circular path on a piece of gauze.
    \item Balloon resection: Carefully incise the outer balloon without puncturing the inner balloon, which is filled with water. 
    \item Suture \& knot: Pass a suture through a Penrose drain and close the slit with a laparoscopic knot consisting of three throws.
\end{itemize}
Among these, peg transfer, circle cutting, and suture \& knot are fundamental training tasks adapted from the MISTELS curriculum. The balloon resection task was developed at the University Hospital Carl Gustav Carus, Dresden, Germany \cite{bechtolsheim2020hunger}. Detailed task instructions are provided in the supplementary material, and representative video frames are shown in Fig.~\ref{fig:training_tasks}.

To capture a broad spectrum of surgical skill, recordings were collected primarily from medical students participating in an elective course on laparoscopic surgery. Each student was recorded at multiple stages of their training: (1) after an introductory lecture and review of instructional videos, (2) after the practical demonstration and guided hands-on training of each task, and finally after each of two independent practice sessions (3, 4). 
During each recording session, referred to as a \emph{trial}, participants completed all four tasks in the following order: peg transfer, circle cutting, balloon resection, and suture \& knot.

\begin{figure}[htb]
\centering

\begin{tikzpicture}
  % Draw horizontal line
  \draw[thick, ->] (0,0) -- (15.5,0);

  % Define timeline points with labels
  \foreach \x/\label in {%
      0/Theoretical\\introduction,
      2/{Peg transfer},
      4.25/{Circle cutting\\+ Balloon resection},
      6.5/{Suture \& knot},
      8.5/Practice,
      10.5/Practice%
      }
      {
        % Training course: filled circles, labels below timeline
        \fill (\x,0) circle (2pt);    
        \node[anchor=north,yshift=-3pt, align=center, font=\small] at (\x,0) {\label};
      }

  \foreach \x/\label in {%
      1/Trial 1,
      7.5/Trial 2,
      9.5/Trial 3,
      11.5/Trial 4,
      13.5/Trial 5,
      14.5/Trial 6%
      }
      {
        % Trials: outlined circles, labels above timeline
        \draw[fill=white] (\x,0) circle (2pt);    
        \node[anchor=north,yshift=15pt, align=center, font=\small] at (\x,0) {\label};
      }

  % add bracket
  \draw [decorate, decoration={brace, mirror, amplitude=10pt}]
    (1.27,-0.9) -- (7.25,-0.9) node[midway, below=10pt, font=\small] {Practical introduction};    
\end{tikzpicture}
\caption{Timeline of video recording during the laparoscopic training course. Filled circles indicate course lessons, while open circles indicate recording sessions.}
\label{fig:timeline}
\end{figure}
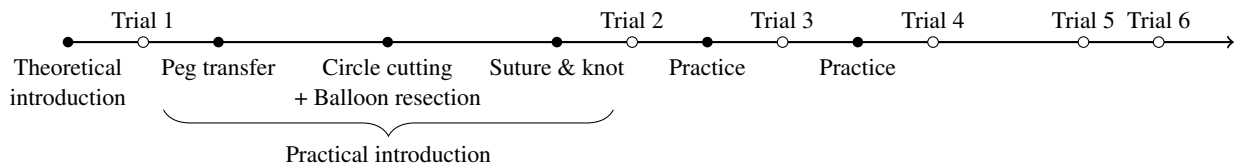

Of all students, fifty participated in an additional study investigating the effect of caffeine intake on laparoscopic performance. In this context, they completed two additional recording sessions (trials 5 and 6) on two separate days after completing the training course. LASANA includes the obtained recordings as examples of advanced skill execution. Notably, caffeine intake was found to have no significant impact on laparoscopic performance \cite{bechtolsheim2022does}. The complete timeline of training and recording sessions is presented in Fig.~\ref{fig:timeline}.

To further enhance the diversity of the dataset, additional recordings were obtained from clinicians with varying levels of laparoscopic experience. Each clinician performed all four tasks once, following the specified task sequence.

In total, 335~stereo video recordings were acquired per task, contributed by 58~medical students and 12~clinicians. The recordings have been analyzed previously in studies investigating laparoscopic skill acquisition (trials 1--4) \cite{bechtolsheim2023does} and the effects of caffeine intake (trials 5 and 6) \cite{bechtolsheim2022does}.
Data collection was approved by the local ethics committee of TU Dresden (EK 416092015), and informed consent was obtained from all participants. 

\subsection*{Video curation and annotation}

Initial preprocessing involved trimming each video to include only the time span from the beginning to the completion of the corresponding task. Video processing was implemented in Python using the OpenCV \cite{bradski2000opencv} and FFmpeg \cite{tomar2006converting} libraries. Further, random pseudo-English identifiers were generated for each recording using the Gibberish library \cite{gibberish}, and all video files were renamed accordingly. As a result, filenames do not contain any information about the participant identity or the timing of the recording session (trial number).

A subset of recordings was removed from the dataset based on predefined \emph{exclusion} criteria. Excluded recordings include those performed under inadequate conditions, such as poor illumination or unsuitable materials (e.g. excessively long sutures for the suture \& knot task, inappropriately coarse gauze for the circle cutting task), as well as recordings of incorrectly executed tasks (e.g. attempting to tie a knot without following the proper laparoscopic technique). 
In addition, videos exceeding a predefined time limit (6\,min for peg transfer, 10\,min for all other tasks) were removed. For the suture \& knot task, specifically, recordings were truncated at 10\,min and retained if at least one throw was completed within this time frame.

All retained recordings were manually annotated with task-specific errors and a structured global skill rating. Annotations were performed by reviewing the video stream from the left camera.

\subsubsection*{Error annotation}

Task-specific errors were defined for each training task. The list below provides the shorthand \texttt{name} of each error and a brief description:
\begin{itemize}[parsep=0pt]
    \item Peg transfer% 
        \begin{itemize}[parsep=0pt]
            \item \texttt{object\_dropped\_within\_fov:} A triangular object is dropped within the visible field and can be retrieved.
            \item \texttt{object\_dropped\_outside\_of\_fov:} A triangular object is dropped outside the visible field or cannot be retrieved.
        \end{itemize}       
    \item Circle cutting%
        \begin{itemize}[parsep=0pt]
            \item \texttt{cutting\_imprecise:} The cutting path deviates from the marked circle.
            \item \texttt{gauze\_detached:} The gauze becomes detached from one or more of the metal clips.
        \end{itemize}
    \item Balloon resection%
        \begin{itemize}[parsep=0pt]
            \item \texttt{cutting\_imprecise:} The cutting path deviates from the marked line.
            \item \texttt{cutting\_incomplete:} The marked line is not fully cut.
            \item \texttt{balloon\_opened:} The inner balloon is opened, meaning that more than half of the contained water leaks out. 
            \item \texttt{balloon\_damaged:} The inner balloon is damaged, but less than half of the contained water leaks out.
            \item \texttt{balloon\_perforated:} The inner balloon is perforated, resulting in only minor leakage (e.g. single drops or leakage under pressure).
        \end{itemize}
    \item Suture \& knot%
        \begin{itemize}[parsep=0pt]
            \item \texttt{needle\_dropped:} The suture needle is dropped.
            \item \texttt{suture\_imprecise:} The suture placement deviates from one or both of the marked dots. % on the Penrose drain.
            \item\texttt{fewer\_than\_three\_throws:} Fewer than three throws are successfully completed.
            \item \texttt{slit\_not\_closed:} The knot does not close the slit in the Penrose drain properly.
            \item \texttt{knot\_comes\_apart:} The knot loosens or comes apart.
            \item \texttt{drain\_detached:} The Penrose drain detaches from the Velcro strip.
        \end{itemize}
\end{itemize}
Errors were annotated at the video level, indicating only whether a given error occurred during the task, without specifying its frequency or timing. Importantly, annotations were based solely on observable events in the video. For example, the strength of the laparoscopic knot was not tested physically after recording an execution of the suture \& knot task. All errors are annotated as binary outcomes (yes/no), even though certain errors (e.g. \texttt{cutting\_imprecise}, \texttt{suture\_imprecise}) could vary in severity. Providing more fine-grained error annotations is planned for future work.

\subsubsection*{Skill annotation}

To evaluate surgical performance, we developed a structured rating tool inspired by the GOALS framework. All GOALS dimensions were adopted except for autonomy, which was deemed irrelevant for the brief, straightforward training tasks. The remaining four assessed dimensions, or skill aspects, are $A := \lbrace{\text{depth perception, efficiency, bimanual dexterity, tissue handling}\rbrace}$.

Each aspect $a \in A$ is rated on a five-point Likert scale, with task-specific anchor descriptors provided in the annotation guidelines (see supplementary material). The overall skill score, referred to as the total \emph{Global Rating Score (GRS)}, is computed as the sum across all four aspects. Formally, the skill rating for a video~$v$ is represented as a collection of scores $p_a(v)$, where $a \in A \cup \lbrace{\text{GRS}\rbrace}$ and $p_{\text{GRS}}(v) := \sum_{{a'} \in A} p_{a'}(v)$.

If a participant clearly failed to complete the task, no skill rating was assigned. Instead, the recording was marked as \emph{failed}. For the balloon resection task, this occurred when the inner balloon was opened with the first cut; for the suture \& knot task, this occurred when fewer than three throws were completed. No failure criteria are applicable to the peg transfer and circle cutting tasks because these are straightforward to complete successfully.

\subsubsection*{Annotation process}

To enhance the reliability of the skill ratings, each video was evaluated independently by three raters, based on which the final scores were obtained by averaging. Averaging multiple independent estimates reduces noise roughly by a factor of $\sqrt{m}$, where $m$ is the number of estimates \cite[p. 261]{kahnemann2021noise}.
Ratings were provided by a pool of four trained raters, denoted as $r_i$ for $0 \leq i \leq 3$.
Two raters ($r_1$ and $r_3$) annotated all available videos, whereas the other raters ($r_0$ and $r_2$) each annotated about half of the videos.
All raters were familiar with the laparoscopic training tasks and received a briefing based on the annotation guideline.

Videos were assigned to raters in batches of 64--69 recordings of the same task. Each batch included recordings with different trial numbers to ensure some variation in skill. Within each batch, the order of videos was randomized differently for each rater to minimize order effects. Because videos were renamed with random identifiers, raters had no access to metadata such as participant pseudonyms or trial numbers. This metadata will also remain unpublished.

For each task, the final aggregated skill rating for each recording is computed as follows:
Let $p_a^i(v)$ denote the score assigned by rater~$r_i$ to video~$v$ for aspect~$a$ and let $V^i$ denote the set of all videos of the task rated by $r_i$. For each aspect $a \in A \cup \lbrace{\text{GRS}\rbrace}$, 
the scores are first \emph{normalized} per rater by subtracting the mean and dividing by the standard deviation: % of that rater’s scores for videos of the given task
\begin{equation}
    \hat{p}_a^i(v) := \frac{p_a^i(v) - \bar{p}_a^i}{s_a^i}, \text{ where } \bar{p}_a^i := \frac{1}{|V^i|} \sum_{v \in V^i} p_a^i(v),~ 
    s_a^i := \sqrt{\frac{1}{|V^i| - 1} \sum_{v \in V^i} (p_a^i(v) - \bar{p}_a^i)}~.
\end{equation}
If rater~$r_i$ marked video~$v$ as failed, the normalized total global rating score $\hat{p}_{\text{GRS}}^i(v)$ is set to -2.5, corresponding to the lowest GRS~value in the dataset after normalization, and the normalized aspect-specific scores $\hat{p}_{a'}^i(v)$ for ${a'} \in A$ are left undefined.
After normalization, the final aggregated score $\bar{p}_a(v)$ for video $v$ and for each aspect $a \in A \cup \lbrace{\text{GRS}\rbrace}$ is the average of the available normalized scores:
\begin{equation}
    \bar{p}_a(v) := \frac{1}{\sum_{i=0}^3 \mathbb{I}(a, i, v)}\sum_{i=0}^3 \mathbb{I}(a, i, v) \hat{p}_a^i(v), \text{ where } \mathbb{I} \text{ is an indicator function and }\mathbb{I}(a, i, v) := 
    \begin{cases}
      0 & \text{if } \hat{p}_a^i(v) \text{ is undefined}\footnotemark\\
      1 & \text{else }
    \end{cases}.
\end{equation}
\footnotetext{
    The normalized score $\hat{p}_a^i(v)$ is \emph{undefined} if rater~$r_i$ did not rate video~$v$ or if  rater~$r_i$ marked video~$v$ as failed and $a \neq \text{GRS}$.
}
If fewer than two independent ratings are available for an aspect~$a$ of video~$v$, which means that $\sum_{i=0}^3 \mathbb{I}(a,i, v) < 2$, the aggregated score $\bar{p}_a(v)$ remains undefined.  

\begin{figure}[htb]
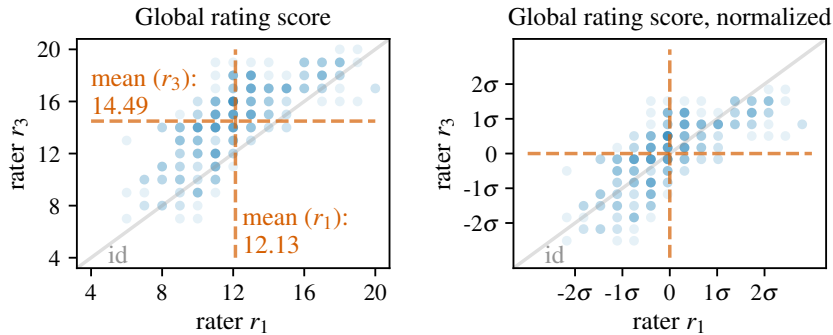

  \centering
  \subcaptionbox{}[0.3\linewidth]{%
    \captionsetup{labelformat=empty,belowskip=-20pt}
    %\frame{
    % trim: left - bottom - right - top
    \begin{adjustbox}{clip,trim=0.25cm 0.05cm 0.6cm 0.1cm,width=\linewidth,keepaspectratio}
        \input{figures/SutureAndKnot_1vs3_normalize-False.pgf}%
    \end{adjustbox}
    %}
  }
  \hspace{1em}
  \subcaptionbox{}[0.3\linewidth]{%
    \captionsetup{labelformat=empty,belowskip=-20pt}
    %\frame{
    % trim: left - bottom - right - top
    \begin{adjustbox}{clip,trim=0.25cm 0.05cm 0.6cm 0.1cm,width=\linewidth,keepaspectratio}
        \input{figures/SutureAndKnot_1vs3_normalize-True.pgf}%
    \end{adjustbox}
    %}
  }
  \caption{Total GRS for suture \& knot recordings assigned by $r_1$ and $r_3$ before (left) and after (right) normalization.}
  \label{fig:normalization}
\end{figure}

Normalization seemed necessary because individual raters naturally tend to use Likert scales differently, which can result in systematic score shifts, or \emph{level noise}, across raters \cite[p. 74]{kahnemann2021noise}. Normalizing scores enhances comparability across raters by aligning their scores to a common standardized scale.
As an illustrative example, Fig.~\ref{fig:normalization} shows the relationship between the total GRS assigned by raters $r_1$ and $r_3$ on the suture \& knot task, before and after rater-wise normalization.

In addition to providing skill ratings, raters also annotated the task-specific errors described previously. These error annotations were subsequently verified and harmonized during a final review round conducted by rater $r_3$.

\subsection*{LASANA benchmark}

\begin{figure}[tb]
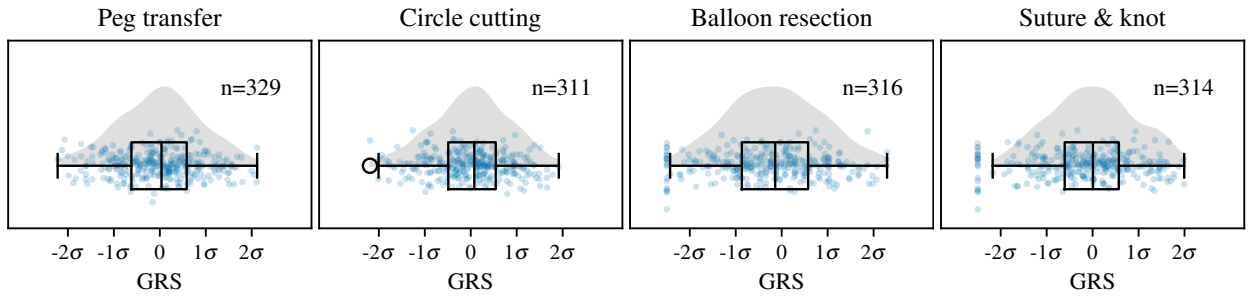

  \centering
  \subcaptionbox{}[0.235\linewidth]{%
    \captionsetup{labelformat=empty,belowskip=-20pt}
    %\frame{
    % trim: left - bottom - right - top
    \begin{adjustbox}{clip,trim=0.64cm 0.1cm 0.56cm 0.04cm,width=\linewidth,keepaspectratio}
        \input{figures/scores_PegTransfer.pgf}%
    \end{adjustbox}
    %}
  }%
  \subcaptionbox{}[0.235\linewidth]{%
    \captionsetup{labelformat=empty,belowskip=-20pt}
    %\frame{
    % trim: left - bottom - right - top
    \begin{adjustbox}{clip,trim=0.64cm 0.1cm 0.56cm 0.04cm,width=\linewidth,keepaspectratio}
        \input{figures/scores_CircleCutting.pgf}%
    \end{adjustbox}
    %}
  }%
  \subcaptionbox{}[0.235\linewidth]{%
    \captionsetup{labelformat=empty,belowskip=-20pt}
    %\frame{
    % trim: left - bottom - right - top
    \begin{adjustbox}{clip,trim=0.64cm 0.1cm 0.56cm 0.04cm,width=\linewidth,keepaspectratio}
        \input{figures/scores_BalloonResection.pgf}%
    \end{adjustbox}
    %}
  }%
  \subcaptionbox{}[0.235\linewidth]{%
    \captionsetup{labelformat=empty,belowskip=-20pt}
    %\frame{
    % trim: left - bottom - right - top
    \begin{adjustbox}{clip,trim=0.64cm 0.1cm 0.56cm 0.04cm,width=\linewidth,keepaspectratio}
        \input{figures/scores_SutureAndKnot.pgf}%
    \end{adjustbox}
    %}
  }
  \caption{Distributions of the total global rating scores per task in the LASANA dataset.}
  \label{fig:scores}
\end{figure}

Figures~\ref{fig:scores} and~\ref{fig:errors} summarize key statistics of the LASANA dataset. The overall skill scores ($\bar{p}_{\text{GRS}}$) % (total global rating scores) 
approximately follow a normal distribution, which reflects the diverse range of surgical skill captured in the dataset. 

Most task-specific errors occur in fewer than one third of the recordings. Some more severe errors (\texttt{object\_dropped\_outside\_of\_fov}, \texttt{fewer\_than\_three\_throws}, \texttt{slit\_not\_closed}, \texttt{knot\_comes\_apart}, \texttt{drain\_detached}) are comparatively rare, appearing in 10\% of the recordings or fewer. This low frequency is not ideal from a benchmarking perspective but needs to be expected because participants were instructed to avoid such critical mistakes.

\begin{figure}[tb]
  \centering
  \subcaptionbox{}[0.235\linewidth]{%
    \captionsetup{labelformat=empty,belowskip=-20pt}
    %\frame{
    % trim: left - bottom - right - top
    \begin{adjustbox}{clip,trim=0.18cm 0.2cm 0.5cm 0.cm,width=\linewidth,keepaspectratio}
        \input{figures/errors_PegTransfer.pgf}%
    \end{adjustbox}
    %}
  }%
  \subcaptionbox{}[0.235\linewidth]{%
    \captionsetup{labelformat=empty,belowskip=-20pt}
    %\frame{
    % trim: left - bottom - right - top
    \begin{adjustbox}{clip,trim=0.18cm 0.2cm 0.5cm 0.cm,width=\linewidth,keepaspectratio}
        \input{figures/errors_CircleCutting.pgf}%
    \end{adjustbox}
    %}
  }%
  \subcaptionbox{}[0.235\linewidth]{%
    \captionsetup{labelformat=empty,belowskip=-20pt}
    %\frame{
    % trim: left - bottom - right - top
    \begin{adjustbox}{clip,trim=0.18cm 0.2cm 0.5cm 0.cm,width=\linewidth,keepaspectratio}
        \input{figures/errors_BalloonResection.pgf}%
    \end{adjustbox}
    %}
  }%
  \subcaptionbox{}[0.235\linewidth]{%
    \captionsetup{labelformat=empty,belowskip=-20pt}
    %\frame{
    % trim: left - bottom - right - top
    \begin{adjustbox}{clip,trim=0.18cm 0.2cm 0.5cm 0.cm,width=\linewidth,keepaspectratio}
        \input{figures/errors_SutureAndKnot.pgf}%
    \end{adjustbox}
    %}
  }
  \caption{Ratio of videos with task-specific errors. The x-axis labels are defined as follows: \small{P0\,=\,\texttt{object\_dropped\_within\_fov}, P1\,=\,\texttt{object\_dropped\_outside\_of\_fov}, C0\,=\,\texttt{cutting\_imprecise}, C1\,=\,\texttt{gauze\_detached}, B0\,=\,\texttt{cutting\_imprecise}, B1\,=\,\texttt{cutting\_incomplete}, B2\,=\,\texttt{balloon\_opened}, B3\,=\,\texttt{balloon\_damaged}, B4\,=\,\texttt{balloon\_perforated}, S0\,=\,\texttt{needle\_dropped}, S1\,=\,\texttt{suture\_imprecise}, S2\,=\,\texttt{fewer\_than\_three\_throws}, S3\,=\,\texttt{slit\_not\_closed}, S4\,=\,\texttt{knot\_comes\_apart}, S5\,=\,\texttt{drain\_detached}}.
  }
  \label{fig:errors}
\end{figure}

To provide a standardized benchmark for the development and evaluation of automatic video-based methods for skill assessment and error recognition, we define a data split for each task. Each split partitions the task recordings into training, validation, and test subsets in an approximate 75:10:15 ratio. The training set should be used for model parameter optimization, while the validation set can be utilized for hyperparameter optimization and monitoring model performance during training. The test set, by contrast, is reserved for the final, unbiased evaluation of any trained model. Table~\ref{tab:datasplit} lists the number of recordings in each subset for all four tasks.

\begin{table}[htb]
    \newcolumntype{g}{>{\columncolor{gray!10}}p{1.5cm}}
    \caption{Number of videos in the LASANA benchmark.}
    \label{tab:datasplit}
    \centering
        \begin{tabular}{p{2.5cm}gp{1.5cm}p{1.5cm}p{1.5cm}}    
        \toprule
        Task & Total & \makecell[cl]{Training \\set} & \makecell[cl]{Validation \\set} & Test set \\
        \midrule
        Peg transfer & 329 & 243 & 32 & 54 \\
        Circle cutting & 311 & 234 & 27 & 50 \\
        Balloon resection & 316 & 235 & 30 & 51 \\
        Suture \& knot & 314 & 232 & 32 & 50 \\
        \bottomrule
    \end{tabular}
\end{table}

Importantly, the split is performed at the participant level, ensuring that all recordings from a single individual are placed entirely within one subset. This guarantees that models are tested on data from previously unseen participants, enabling a realistic assessment of generalization ability. The split is sampled randomly, where the distributions of the total GRS and the error frequencies are balanced across the subsets. Additionally, videos with strong inter-rater disagreement on the skill rating\footnote{
We quantify the disagreement on the skill rating for a video~$v$ by computing the mean absolute deviation $\epsilon(v)$ between the aggregated normalized total GRS and each individual rater’s normalized total GRS. More specifically, $
\epsilon(v) = \frac{1}{\sum_{i=0}^3 \mathbb{I}(\text{GRS}, i, v)}\sum_{i=0}^3 \mathbb{I}(\text{GRS}, i, v) \, |\hat{p}_{\text{GRS}}^i(v) - \bar{p}_{\text{GRS}}(v)|~.
$
A video~$v$ has strong disagreement on its skill rating if $\epsilon(v)$ exceeds the 95th percentile of $\lbrace{\epsilon(v'): v' \in V \rbrace}$, where $V$ denotes all videos of the corresponding task.
} are guaranteed to be placed into the training set instead of the validation and test sets.

\section*{Data Records}

The LASANA dataset \cite{funke2026lasana} comprises over 310 trimmed and synchronized stereo video recordings for each of the four laparoscopic training tasks (see Table~\ref{tab:datasplit}). Each recording includes binary error annotations at the video level and a skill rating aggregated from three independent raters.
Each recording is assigned a unique pseudo-English identifier, written in lowercase letters. 

\begin{figure}[htb]
    \centering
    %\frame{
    % trim: left - bottom - right - top
    \begin{adjustbox}{clip,trim=0.cm 0.2cm 0.2cm 0.2cm,width=0.95\linewidth,keepaspectratio}
        \includegraphics{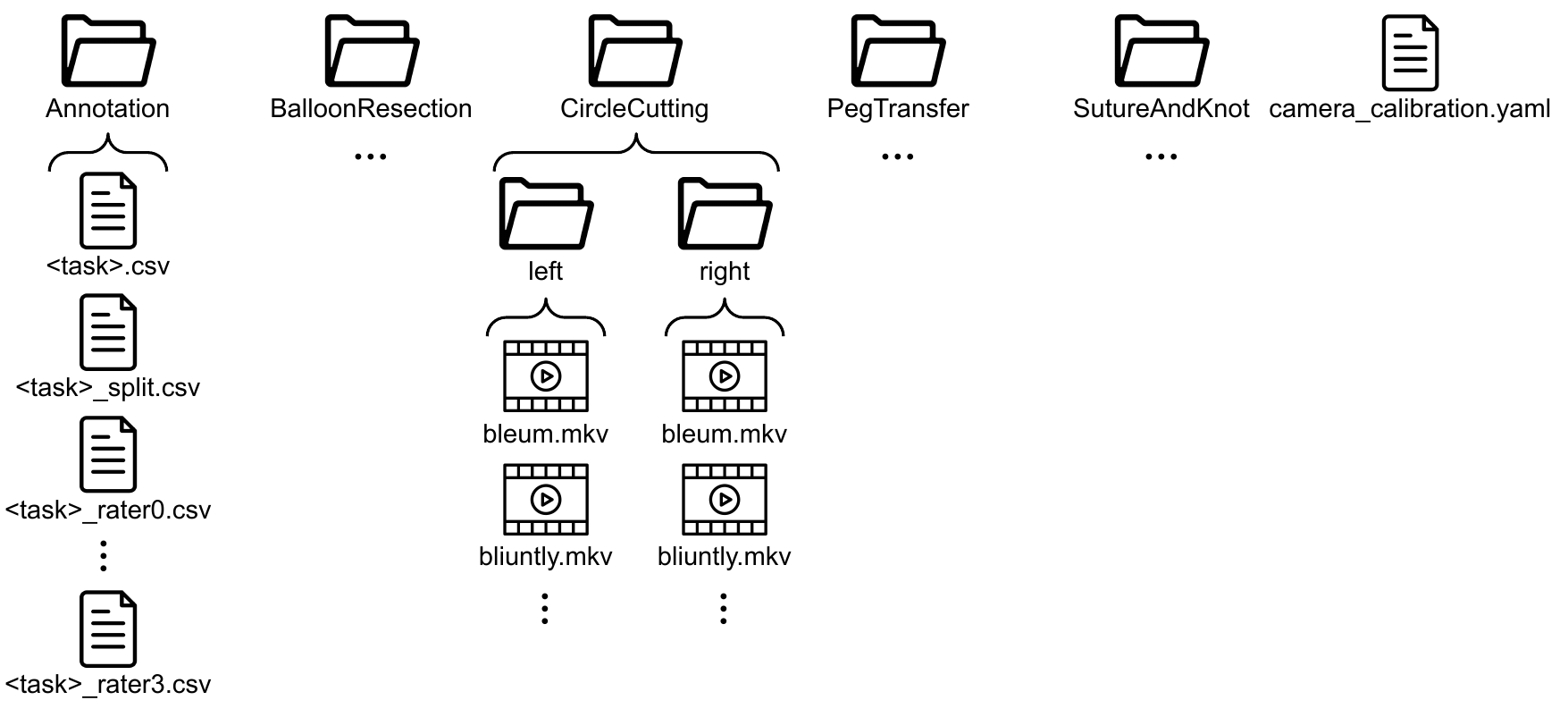}%
    \end{adjustbox}
    %}   
    \caption{    
    The folder and file structure of the LASANA dataset. Each task has its own corresponding set of annotation files, where \texttt{<task>} is replaced with the task name written in UpperCamelCase.
    }
    \label{fig:folder_structure}
\end{figure}

The overall file structure is shown in Fig.~\ref{fig:folder_structure}.
The recordings are organized by task.  
Folders named after each task, written in \emph{UpperCamelCase}, store the corresponding stereo videos. 
Here, the video streams from the left and right cameras are stored as separate files in the respective subfolders, \texttt{left} and \texttt{right}, where each video file is named after the recording’s identifier. 

To enable full use of the stereo video information, the dataset includes camera calibration details in the file \texttt{camera\_calibration.yaml}, formatted in YAML syntax. Specifically, this file provides the intrinsic parameters of both the left and right cameras, as well as their relative pose (rotation and translation). We provide sample Python code demonstrating how to perform image rectification and stereo matching using the supplied calibration data, based on OpenCV and Unimatch \cite{xu2023unifying}.

Annotation files for all tasks are located in the folder \texttt{Annotation}. These are provided as CSV files with semicolon delimiter. Each CSV file contains one row per recording, where the first column (named \texttt{id}) specifies the recording's identifier.

For each task, there is a main annotation file named \texttt{<task>.csv}, where \texttt{<task>} is replaced with the task name in UpperCamelCase. The main annotation file comprises:
\begin{itemize}[parsep=0pt]
    \item A column for each task-specific error, which contains \texttt{True} if the error (as defined in the column header) occurred at least once during the recording, and \texttt{False} otherwise.
    \item A column named \texttt{GRS}, which contains the aggregated normalized total skill score $\bar{p}_\text{GRS}$ of the recording.
    \item Additional columns named after each rating aspect $a' \in \lbrace{\text{depth perception, efficiency, bimanual dexterity, tissue handling}\rbrace}$, written in \emph{snake\_case}. These columns store the aggregated normalized scores $\bar{p}_{a'}$ for the corresponding aspects. Missing values in these columns occur for recordings that were marked as failed by more than one rater.
    \item A column named \texttt{duration}, which specifies the length of the video recording in \texttt{0:mm:ss} format.
    \item A column named \texttt{frame\_count}, which indicates the total number of frames in both the left and the right video.
\end{itemize}

In addition, the dataset includes the original scores assigned by each rater $r_i$, where $0 \leq i \leq 3$. For each task and rater, these scores are stored in file \texttt{<task>\_rater\,$i$.csv}. These files contain a column \texttt{rater} indicating the rater’s identifier~$i$ as well as one column for each aspect $a \in \{$depth perception, efficiency, bimanual dexterity, tissue handling, GRS$\}$ containing the assigned score~$p_a^i$. If a recording was marked as failed by the rater, the rating scores are undefined, indicated by missing entries.

Finally, the data split for each task is provided in file \texttt{<task>\_split.csv}, where the column \texttt{split} specifies the subset assignment (\texttt{train}, \texttt{val}, or \texttt{test}) of each recording.

The LASANA dataset can be accessed at \url{https://doi.org/10.25532/OPARA-1046}. For each task, the folders \texttt{<task>/left} and \texttt{<task>/right} are stored as zip files \texttt{<task>\_left.zip} and \texttt{<task>\_right.zip}. The \texttt{Annotation} folder is also stored as \texttt{Annotation.zip}. In addition, the file \texttt{example\_videos.zip} provides one example video for each task, recorded with the left camera.

\section*{Technical Validation}

\subsection*{Statistical analysis}

\begin{figure}[htb]
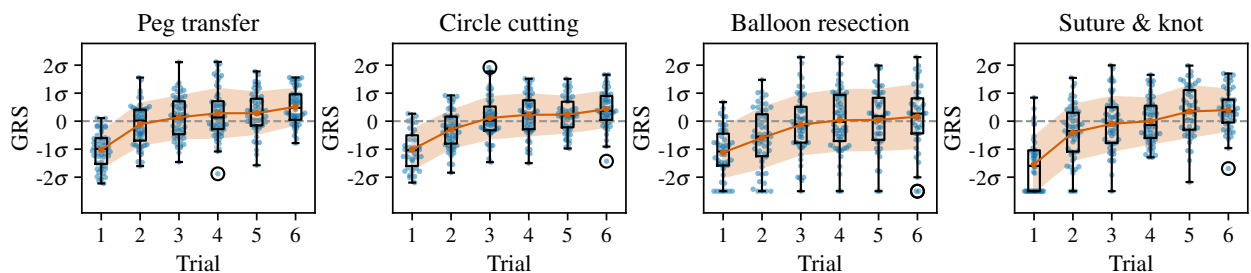

  \centering
  \subcaptionbox{}[0.235\linewidth]{%
    \captionsetup{labelformat=empty,belowskip=-20pt}
    %\frame{
    % trim: left - bottom - right - top
    \begin{adjustbox}{clip,trim=0.2cm 0.05cm 0.5cm 0.02cm,width=\linewidth,keepaspectratio}
        \input{figures/PegTransfer_devel.pgf}%
    \end{adjustbox}
    %}
  }%
  \subcaptionbox{}[0.235\linewidth]{%
    \captionsetup{labelformat=empty,belowskip=-20pt}
    %\frame{
    % trim: left - bottom - right - top
    \begin{adjustbox}{clip,trim=0.2cm 0.05cm 0.5cm 0.03cm,width=\linewidth,keepaspectratio}
        \input{figures/CircleCutting_devel.pgf}%
    \end{adjustbox}
    %}
  }%
  \subcaptionbox{}[0.235\linewidth]{%
    \captionsetup{labelformat=empty,belowskip=-20pt}
    %\frame{
    % trim: left - bottom - right - top
    \begin{adjustbox}{clip,trim=0.2cm 0.05cm 0.5cm 0.03cm,width=\linewidth,keepaspectratio}
        \input{figures/BalloonResection_devel.pgf}%
    \end{adjustbox}
    %}
  }%
  \subcaptionbox{}[0.235\linewidth]{%
    \captionsetup{labelformat=empty,belowskip=-20pt}
    %\frame{
    % trim: left - bottom - right - top
    \begin{adjustbox}{clip,trim=0.2cm 0.05cm 0.5cm 0.03cm,width=\linewidth,keepaspectratio}
        \input{figures/SutureAndKnot_devel.pgf}%
    \end{adjustbox}
    %}
  }
  \caption{Development of total global rating scores across consecutive recording sessions. Small dots indicate scores for individual recordings, summarized using box plots. For each trial, the solid orange dot marks the mean and the shaded orange area represents the standard deviation of the scores corresponding to that trial.}
  \label{fig:development}
\end{figure}

To evaluate the validity of the skill ratings provided in the dataset, we analyze whether the total GRS can effectively distinguish between participants at different stages of skill acquisition. For this purpose, the recordings of each task are grouped according to their trial number, and the corresponding GRS values are shown in Fig.~\ref{fig:development}.
As anticipated, the GRS increases progressively over time, indicating skill improvement through practice, although the learning curve begins to plateau after the third trial. 

\begin{figure}[htb]
  \centering
    %\frame{
    %trim: left - bottom - right - top
    \begin{adjustbox}{clip,trim=0.18cm 0.0cm 0.67cm 0.1cm,width=0.28\linewidth,keepaspectratio}
        \input{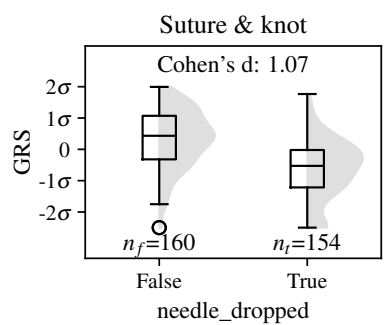}%
    \end{adjustbox}
    %}
  \caption{Impact of needle drops in the suture \& knot task. Recordings in which the needle was dropped % at least once 
  systematically receive lower total global rating scores.}
  \label{fig:error_effect}
\end{figure}

We further examine the relationship between annotated task-specific errors and the overall skill score. It is expected that the group~$T$ of recordings in which a specific error occurred would, on average, receive a lower GRS compared to the group~$F$ without that error.
An example of this relation is illustrated in Fig.~\ref{fig:error_effect}, which compares total global rating scores for suture \& knot recordings depending on whether the needle was dropped or not.

To quantify the influence of each error on the total GRS, we compute the effect size using Cohen's~$d$ \cite[p. 66f]{cohen1988statistical}, which represents the standardized difference between the means of two groups\footnote{
Given sample data $X:=(x_i)_{i=1}^{n_x}$, the mean~$\bar{x}$, variance~$s_x^2$, and standard deviation~$s_x$ are defined as
$
    \bar{x} := \frac{1}{n_x} \sum_{i=1}^{n_x} x_i,~ 
    s_x^2 := \frac{1}{n_x - 1} \sum_{i=1}^{n_x} (x_i - \bar{x}),~s_x := \sqrt{s_x^2}~.
    % \text{ If possible, we write $n$ instead of $n_x$ for simplicity.}
$
}
:
\begin{equation}
    d := \frac{\bar{f} - \bar{t}}{s}, \text{ where } s := \sqrt{\frac{(n_f - 1)s_f^2 + (n_t - 1)s_t^2}{n_f + n_t - 2}} \text{ denotes the pooled standard deviation.}
\end{equation}

The effect sizes for all task-specific errors are reported in Table~\ref{tab:error_effect}. As expected, most errors exhibit a medium to large effect on the GRS.
Errors that inherently cause or are likely to cause task failure (e.g. \texttt{balloon\_opened} and \texttt{fewer\_than\_three\_throws}) have the strongest effects.
In contrast, the errors defined for the circle cutting task have only medium effects, while on the suture \& knot task, errors related to suture accuracy and tightness show only small effects on the GRS.

\begin{table}[tb]
    % measure size of box
    \setbox9=\hbox{
        \subcaptionbox{}[0.32\linewidth]{%
            \captionsetup{labelformat=empty,belowskip=-20pt}
        
            \begin{adjustbox}{width=\linewidth}
        
            \begin{tabular}{lc}    
                \multicolumn{2}{c}{Peg transfer} \\
                \toprule
                Error & $d$ \\
                \midrule
                \texttt{object\_dropped\_within\_fov} & \cellcolor{lightblue} 0.96 \\
                \texttt{object\_dropped\_outside\_of\_fov} & \cellcolor{lightblue} 1.08 \\
                \bottomrule
            \end{tabular}
        
            \end{adjustbox}
        
            \vspace{0.5em}
        
            \begin{adjustbox}{width=0.68\linewidth}
        
            \begin{tabular}{lc}    
                \multicolumn{2}{c}{Circle cutting} \\
                \toprule
                Error & $d$ \\
                \midrule
                \texttt{cutting\_imprecise} & \cellcolor{lightblue!40} 0.55 \\
                \texttt{gauze\_detached} & \cellcolor{lightblue!40} 0.56 \\
                \bottomrule
            \end{tabular}   
            \end{adjustbox}
          }
    }

  \centering
  \caption{Effect of task-specific errors on the total GRS, quantified by Cohen's $d$. All errors show at least a small effect ($|d| > 0.2$). Cells with \colorbox{lightblue!40}{$|d| > 0.5$} (medium effect size) and \colorbox{lightblue}{$|d| > 0.8$} (large effect size) are highlighted.}
  \label{tab:error_effect}    
  
  \subcaptionbox{}[0.32\linewidth]{%
    \captionsetup{labelformat=empty,belowskip=-20pt}

    \begin{adjustbox}{width=\linewidth}

    \begin{tabular}{lc}    
        \multicolumn{2}{c}{Peg transfer} \\
        \toprule
        Error & $d$ \\
        \midrule
        \texttt{object\_dropped\_within\_fov} & \cellcolor{lightblue} 0.96 \\
        \texttt{object\_dropped\_outside\_of\_fov} & \cellcolor{lightblue} 1.08 \\
        \bottomrule
    \end{tabular}

    \end{adjustbox}

    \vspace{0.5em}

    \begin{adjustbox}{width=0.68\linewidth}

    \begin{tabular}{lc}    
        \multicolumn{2}{c}{Circle cutting} \\
        \toprule
        Error & $d$ \\
        \midrule
        \texttt{cutting\_imprecise} & \cellcolor{lightblue!40} 0.55 \\
        \texttt{gauze\_detached} & \cellcolor{lightblue!40} 0.56 \\
        \bottomrule
    \end{tabular}   
    \end{adjustbox}
  }
  \hspace{1em}
  \subcaptionbox{}[0.235\linewidth]{%
    \captionsetup{labelformat=empty,belowskip=-20pt}

    \raisebox{\dimexpr\ht9-\height}{

    \begin{adjustbox}{width=\linewidth,valign=t}
    
    \begin{tabular}{lc}    
        \multicolumn{2}{c}{Balloon resection} \\
        \toprule
        Error & $d$ \\
        \midrule
        \texttt{cutting\_imprecise} & \cellcolor{lightblue!40} 0.75 \\
        \texttt{cutting\_incomplete} & \cellcolor{lightblue!40} 0.56 \\
        \texttt{balloon\_opened} & \cellcolor{lightblue} 2.05 \\
        \texttt{balloon\_damaged} & \cellcolor{lightblue} 1.27 \\
        \texttt{balloon\_perforated} & \cellcolor{lightblue!40} 0.63  \\
        \bottomrule
    \end{tabular}

    \end{adjustbox}

    }
  }
  \hspace{1em}
  \subcaptionbox{}[0.27\linewidth]{%
    \captionsetup{labelformat=empty,belowskip=-20pt}

    \raisebox{\dimexpr\ht9-\height}{

    \begin{adjustbox}{width=\linewidth,valign=t}
    
    \begin{tabular}{lc}    
        \multicolumn{2}{c}{Suture \& knot} \\
        \toprule
        Error & $d$ \\
        \midrule
        \texttt{needle\_dropped} & \cellcolor{lightblue} 1.07 \\
        \texttt{suture\_imprecise} & \cellcolor{white} 0.23 \\
        \texttt{fewer\_than\_three\_throws} & \cellcolor{lightblue} 2.86 \\
        \texttt{slit\_not\_closed} & \cellcolor{white} 0.37 \\
        \texttt{knot\_comes\_apart} & \cellcolor{lightblue} 1.13 \\
        \texttt{drain\_detached} & \cellcolor{lightblue!40} 0.57 \\
        \bottomrule
    \end{tabular}

    \end{adjustbox}
    }
  }
\end{table}

Finally, we examine the agreement between raters on the skill ratings.
To quantify the agreement between raters $r_x$ and $r_y$, we use \emph{Lin's Concordance Correlation Coefficient}~$\rho_c$ \cite{lawrence1989concordance}.
Let $X:=(x_i)_{i=1}^{n}$ and $Y:=(y_i)_{i=1}^{n}$ denote paired vectors of normalized skill scores, where for $1 \leq i \leq n$, $x_i$ is the score assigned by rater~$r_x$ and $y_i$ is the score assigned by rater~$r_y$ for the $i$-th recording of a task. 
Similar to Cohen's~$\kappa$~\cite{cohen1960coefficient}, Lin's~$\rho_c$ quantifies agreement by relating the observed disagreement to the disagreement that can be attributed to chance\footnote{
    Given paired sample data $X:=(x_i)_{i=1}^{n}$ and $Y:=(y_i)_{i=1}^{n}$, where each pair $(x_i, y_i)$ corresponds to two observations on the same individual, the covariance~$s_{xy}$ and Pearson's correlation coefficient~$r(X, Y)$ to measure the linear relationship between $X$ and $Y$ are defined as 
    $
        s_{xy} := \frac{1}{n - 1} \sum_{i=1}^n (x_i - \bar{x}) (y_i - \bar{y}), ~r(X, Y) := \frac{s_{xy}}{s_x s_y}~.
    $
}:
\begin{equation} \label{eq:ccc}
\begin{split}
\rho_c(X, Y) :=&~ 1 - \frac{\text{mean squared difference between X and Y}}{\text{expected squared difference for uncorrelated X and Y}} = 1 - \frac{\frac{1}{n}\sum_{i=1}^n(x_i - y_i)^2}{s_x^{2} + s_y^{2} + (\bar{x} - \bar{y})^2} \\ =&~ \frac{2 s_{xy}}{s_x^{2} + s_y^{2} + (\bar{x} - \bar{y})^2} = r(X,Y) \frac{1}{\frac{1}{2} (\frac{s_x}{s_y} + \frac{s_y}{s_x} + \frac{(\bar{x} - \bar{y})^2}{s_x s_y})}~.
\end{split}
\end{equation}
Equation~(\ref{eq:ccc}) shows that Lin’s $\rho_c(X, Y)$ can be interpreted in two ways: (1) as a measure of the mean squared error between $X$ and $Y$, transformed to be bounded between -1 and 1\footnote{
This follows from $-(s_x^2 + s_y^2)\leq 2s_{xy} \leq s_x^2 + s_y^2$, which implies $-\frac{s_x^2 + s_y^2}{s_x^{2} + s_y^{2} + (\bar{x} - \bar{y})^2} \leq\rho_c(X,Y) \leq \frac{s_x^2 + s_y^2}{s_x^{2} + s_y^{2} + (\bar{x} - \bar{y})^2}$, where $(\bar{x} - \bar{y})^2 > 0$.
}; and (2) as a scaled version of Pearson’s correlation coefficient~$r(X,Y)$, adjusted to account for differences in scale (relation between $s_x$ and $s_y$) and location (distance between $\bar{x}$ and $\bar{y}$).
In the case of perfect agreement ($x_i = y_i$ for all $i$), $\rho_c(X, Y) = 1$. For perfect inverse agreement ($x_i = - y_i$ for all $i$), $\rho_c(X, Y) = \text{-}1$. If there is no association between $X$ and $Y$, i.e. the covariance $s_{xy} = 0$, then $\rho_c(X,Y) = 0$.

\begin{figure}[htb]
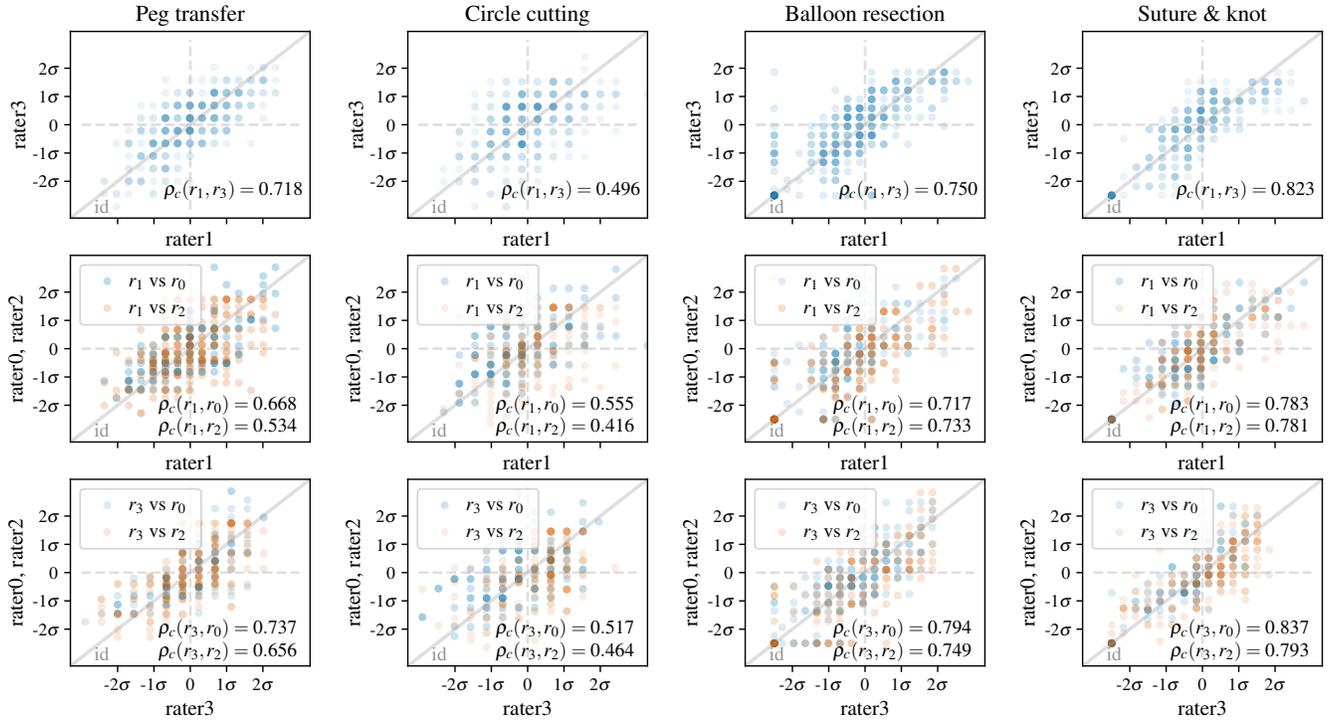

  \centering
  \subcaptionbox{}[0.235\linewidth]{%
    \captionsetup{labelformat=empty,belowskip=-20pt}
    %\frame{
    % trim: left - bottom - right - top
    \begin{adjustbox}{clip,trim=0.13cm 0.5cm 0.5cm 0.2cm,width=\linewidth,keepaspectratio}
        \input{figures/agreement_PegTransfer.pgf}%
    \end{adjustbox}
    %}
  }
  \hfill
  \subcaptionbox{}[0.235\linewidth]{%
    \captionsetup{labelformat=empty,belowskip=-20pt}
    %\frame{
    % trim: left - bottom - right - top
    \begin{adjustbox}{clip,trim=0.13cm 0.5cm 0.5cm 0.2cm,width=\linewidth,keepaspectratio}
        \input{figures/agreement_CircleCutting.pgf}%
    \end{adjustbox}
    %}
  }
  \hfill
  \subcaptionbox{}[0.235\linewidth]{%
    \captionsetup{labelformat=empty,belowskip=-20pt}
    %\frame{
    % trim: left - bottom - right - top
    \begin{adjustbox}{clip,trim=0.13cm 0.5cm 0.5cm 0.2cm,width=\linewidth,keepaspectratio}
        \input{figures/agreement_BalloonResection.pgf}%
    \end{adjustbox}
    %}
  }
  \hfill
  \subcaptionbox{}[0.235\linewidth]{%
    \captionsetup{labelformat=empty,belowskip=-20pt}
    %\frame{
    % trim: left - bottom - right - top
    \begin{adjustbox}{clip,trim=0.13cm 0.5cm 0.5cm 0.2cm,width=\linewidth,keepaspectratio}
        \input{figures/agreement_SutureAndKnot.pgf}%
    \end{adjustbox}
    %}
  }
  \caption{Normalized total global rating scores assigned by different rater pairs for each task. Here, $\rho_c(r_x, r_y)$ denotes the agreement (Lin's Concordance Correlation Coefficient) between the scores assigned by rater~$r_x$ and rater~$r_y$. Because raters $r_0$ and $r_2$ annotated two complementary halves of the videos, their scores are displayed together.}
  \label{fig:agreement}
\end{figure}

The scatter plots in Fig.~\ref{fig:agreement} display the paired normalized GRS values for all pairs of raters and each task. Table~\ref{tab:agreement} summarizes the inter-rater agreement, expressed as the average $\rho_c$ across all rater pairs, for both the total GRS and for each individual skill aspect $a' \in A$.

\begin{table}[htb]
    \newcolumntype{g}{>{\columncolor{gray!10}}p{1.5cm}}
    \caption{Inter-rater agreement on the skill ratings. Average pairwise Lin's $\rho_c$ is shown for the total global rating score and for each individual skill aspect.}
    \label{tab:agreement}
    \centering
        \begin{tabular}{p{2.7cm}gp{1.5cm}p{1.5cm}p{1.5cm}p{1.5cm}}    
        \toprule
        Task & GRS & \makecell[cl]{Depth \\perception} & Efficiency & \makecell[cl]{Bimanual \\dexterity} & \makecell[cl]{Tissue \\handling} \\
        \midrule
        Peg transfer & 0.662 & 0.552 & 0.607 & 0.474 & 0.437 \\
        Circle cutting  & 0.490 & 0.472 & 0.468 & 0.257 & 0.342 \\
        Balloon resection~~~ & 0.749 & 0.516 & 0.565 & 0.464 & 0.648 \\
        Suture \& knot & 0.803 & 0.578 & 0.644 & 0.507 & 0.386 \\
        \bottomrule
    \end{tabular}
\end{table}

As expected for human judgments, there is notable variability among the GRS values that raters assigned  (Fig.~\ref{fig:agreement}). Nevertheless, all tasks except circle cutting achieve an inter-rater agreement of $\rho_c > 0.6$ on the total GRS, which indicates acceptable reliability.
The circle cutting task, however, shows only moderate agreement ($\rho_c > 0.4$). This may be because recordings display only subtle variations in skill or because recordings are difficult to rank consistently when opposing factors are observed. In particular, participants tended to cut with lower precision when they cut with higher efficiency.

Moreover, agreement on individual aspects, or test items, of the rating is generally lower than that for the total score. Because the total GRS aggregates scores over multiple aspects, random variations tend to cancel out, improving consistency.
Enhancing the reliability of aspect-level ratings, potentially through a hierarchical structured assessment framework with multiple test items per aspect, is an important direction for future work.

\begin{figure}[htb]
  \centering
    %\frame{
    %trim: left - bottom - right - top
    \begin{adjustbox}{clip,trim=0.1cm 0.cm 0.15cm 0.1cm,width=0.65\linewidth,keepaspectratio}
        \includegraphics[]{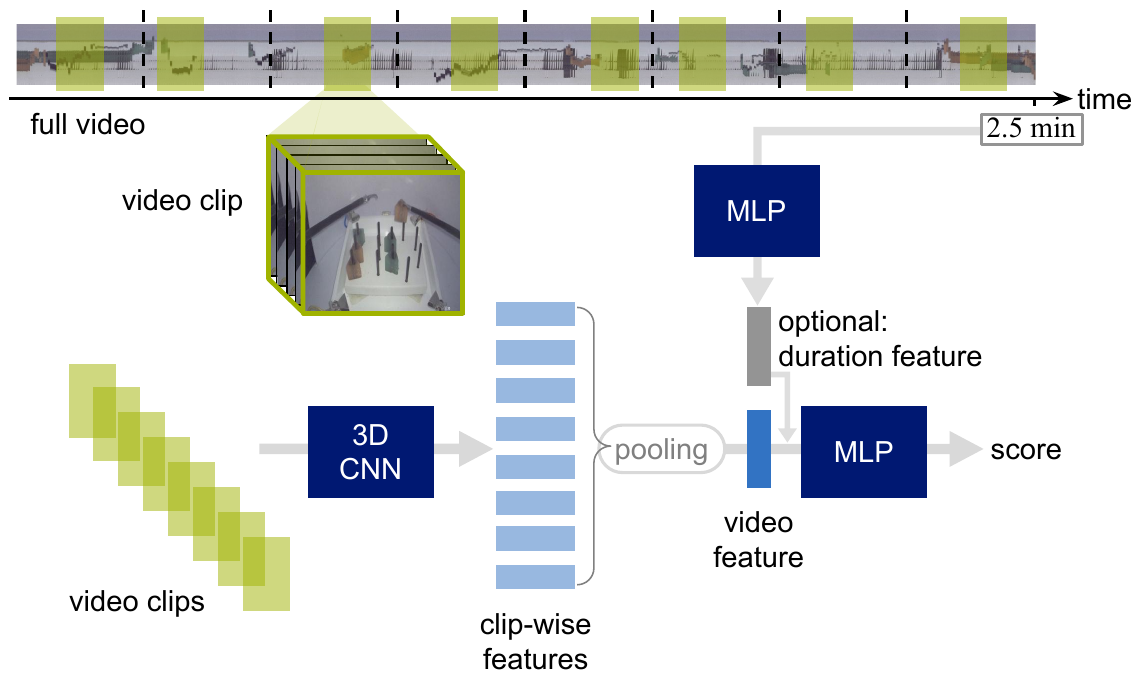}
    \end{adjustbox}
    %}
  \caption{Baseline model for automatic video-based skill score regression.}
  \label{fig:baseline_model}
\end{figure}

\subsection*{Benchmark baselines}

\subsubsection*{Skill score regression}

To encourage reuse of the LASANA dataset for benchmarking automatic video-based skill assessment algorithms, we provide a baseline model for estimating the total GRS for each task. This model leverages an established deep learning architecture for surgical skill assessment \cite{funke2019video}, previously demonstrated to distinguish between novice, intermediate, and expert skill levels on the JIGSAWS dataset. Evaluating the baseline also offers insight into how well an automatic computer vision algorithm can reproduce human ratings, providing additional evidence for the validity of the annotated skill ratings.

The baseline is trained and evaluated using the designated data splits. We suggest to quantify model performance on the test set by measuring agreement with the annotated total GRS using Lin's Concordance Correlation Coefficient~$\rho_c$, see equation~(\ref{eq:ccc}). This metric captures both, linear association between annotated and computed scores as well as shifts in location and scale.   

The baseline model uses the video of the left camera, ignoring stereo information. Input videos are divided into short clips, and a 3D~CNN (X3D \cite{feichtenhofer2020x3d}) extracts features for each clip. The clip-wise features are aggregated via temporal global average pooling to obtain a video-level representation, which is then input to a two-layer MLP to regress the GRS.
Optionally, task completion time can be incorporated by embedding the video duration with a second MLP. This duration feature is concatenated with the video-level representation before score regression. Fig.~\ref{fig:baseline_model} illustrates the model architecture.
Implementation and training details are described in a separate section below.

\begin{figure}[htb]
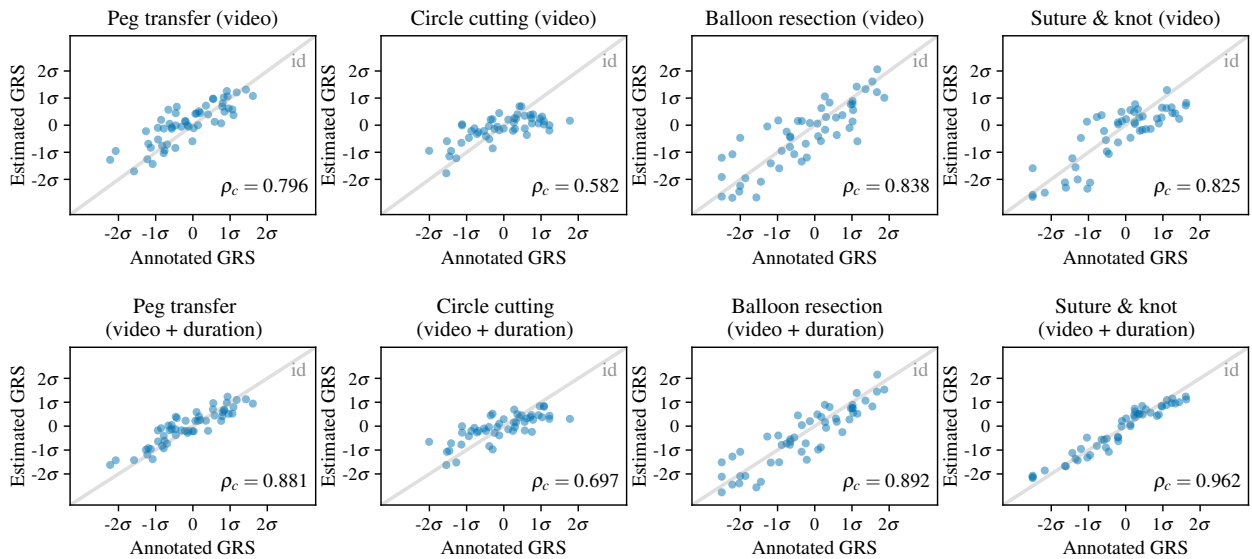

  \centering
    \subcaptionbox{}[0.235\linewidth]{%
    \captionsetup{labelformat=empty,belowskip=-20pt}
    %\frame{
    % trim: left - bottom - right - top
    \begin{adjustbox}{clip,trim=0.18cm 0.03cm 0.6cm 0.1cm,width=\linewidth,keepaspectratio}
        \input{figures/pred_PegTransfer_VideoOnly.pgf}%
    \end{adjustbox}
    %}
  }%
  \subcaptionbox{}[0.235\linewidth]{%
    \captionsetup{labelformat=empty,belowskip=-20pt}
    %\frame{
    % trim: left - bottom - right - top
    \begin{adjustbox}{clip,trim=0.18cm 0.03cm 0.6cm 0.1,width=\linewidth,keepaspectratio}
        \input{figures/pred_CircleCutting_VideoOnly.pgf}%
    \end{adjustbox}
    %}
  }%
  \subcaptionbox{}[0.235\linewidth]{%
    \captionsetup{labelformat=empty,belowskip=-20pt}
    %\frame{
    % trim: left - bottom - right - top
    \begin{adjustbox}{clip,trim=0.18cm 0.03cm 0.6cm 0.1,width=\linewidth,keepaspectratio}
        \input{figures/pred_BalloonResection_VideoOnly.pgf}%
    \end{adjustbox}
    %}
  }%
  \subcaptionbox{}[0.235\linewidth]{%
    \captionsetup{labelformat=empty,belowskip=-20pt}
    %\frame{
    % trim: left - bottom - right - top
    \begin{adjustbox}{clip,trim=0.18cm 0.03cm 0.6cm 0.1,width=\linewidth,keepaspectratio}
        \input{figures/pred_SutureAndKnot_VideoOnly.pgf}%
    \end{adjustbox}
    %}
  }

  \vspace{1em}
  
  \subcaptionbox{}[0.235\linewidth]{%
    \captionsetup{labelformat=empty,belowskip=-20pt}
    %\frame{
    % trim: left - bottom - right - top
    \begin{adjustbox}{clip,trim=0.18cm 0.03cm 0.6cm 0.cm,width=\linewidth,keepaspectratio}
        \input{figures/pred_PegTransfer_VideoDuration.pgf}%
    \end{adjustbox}
    %}
  }%
  \subcaptionbox{}[0.235\linewidth]{%
    \captionsetup{labelformat=empty,belowskip=-20pt}
    %\frame{
    % trim: left - bottom - right - top
    \begin{adjustbox}{clip,trim=0.18cm 0.03cm 0.6cm 0.,width=\linewidth,keepaspectratio}
        \input{figures/pred_CircleCutting_VideoDuration.pgf}%
    \end{adjustbox}
    %}
  }%
  \subcaptionbox{}[0.235\linewidth]{%
    \captionsetup{labelformat=empty,belowskip=-20pt}
    %\frame{
    % trim: left - bottom - right - top
    \begin{adjustbox}{clip,trim=0.18cm 0.03cm 0.6cm 0.,width=\linewidth,keepaspectratio}
        \input{figures/pred_BalloonResection_VideoDuration.pgf}%
    \end{adjustbox}
    %}
  }%
  \subcaptionbox{}[0.235\linewidth]{%
    \captionsetup{labelformat=empty,belowskip=-20pt}
    %\frame{
    % trim: left - bottom - right - top
    \begin{adjustbox}{clip,trim=0.18cm 0.03cm 0.6cm 0.,width=\linewidth,keepaspectratio}
        \input{figures/pred_SutureAndKnot_VideoDuration.pgf}%
    \end{adjustbox}
    %}
  }
  \caption{
  Comparison between annotated and estimated total global rating scores, computed using the model ensembles on the test sets. The upper plots show results for models using video input only while the lower plots correspond to models incorporating both video and duration information.
  }
  \label{fig:results}
\end{figure}

Model training is stochastic, so variability is expected between runs. To consider this variability, training is repeated three times with identical hyperparameters but different random seeds. Then, model performance is quantified by computing mean and standard deviation of Lin's $\rho_c$ across repeated runs. Additionally, the three replicated models are combined in an ensemble, which averages estimates across the individual models, producing potentially more robust ratings. Table~\ref{tab:automatic_assessment} shows results on the validation and test sets for models considering video only or both video and duration as input. For the test videos, Fig.~\ref{fig:results} plots the estimated GRS, computed with the model ensemble, against the annotated GRS.

\begin{table}[htb]
    \caption{Performance of the baseline models for automatic GRS regression on the LASANA validation and test sets, evaluated using Lin’s~$\rho_c$. Reported values denote the mean and standard deviation over three independent runs or the result obtained after \colorbox{gray!10}{ensembling,} which corresponds to averaging GRS estimates across all runs.
    }
    \label{tab:automatic_assessment}
    \centering
    \begin{adjustbox}{width=\linewidth}
        \begin{tabular}{llllllllll}
            \toprule
            \multirow{2}{*}{Input} & \multirow{2}{*}{\makecell[cl]{Ensem-\\ble?}} & \multicolumn{2}{l}{Peg transfer} & \multicolumn{2}{l}{Circle cutting} & \multicolumn{2}{l}{Balloon resection} & \multicolumn{2}{l}{Suture \& knot} \\
             &   & validation & test & validation & test & validation & test & validation & test \\ 
             \midrule
             \multirow{2}{*}{video} & -- & $0.800_{\pm 0.017}$ & $0.782_{\pm 0.040}$ & $0.594_{\pm 0.029}$ & $0.557_{\pm 0.017}$ & $0.800_{\pm 0.010}$ & $0.821_{\pm 0.031}$ & $0.691_{\pm 0.073}$ & $0.803_{\pm 0.015}$ \\
               & \cellcolor{gray!10}\checkmark & \cellcolor{gray!10}$0.814$ & \cellcolor{gray!10}$0.796$ & \cellcolor{gray!10}0.624 & \cellcolor{gray!10}0.582 & \cellcolor{gray!10}0.817 & \cellcolor{gray!10}0.838 & \cellcolor{gray!10}0.709 & \cellcolor{gray!10}0.825 \\
              \midrule
             \multirow{2}{*}{\makecell[cl]{video\\+ duration}} & -- & $0.893_{\pm 0.015}$ & $0.863_{\pm 0.033}$ & $0.783_{\pm 0.033}$ & $0.666_{\pm 0.054}$ & $0.818_{\pm 0.009}$ & $0.879_{\pm 0.007}$ & $0.883_{\pm 0.024}$ & $0.953_{\pm 0.008}$\\
              & \cellcolor{gray!10}\checkmark & \cellcolor{gray!10}0.912 & \cellcolor{gray!10}0.881 & \cellcolor{gray!10}0.813 & \cellcolor{gray!10}0.697 & \cellcolor{gray!10}0.833 & \cellcolor{gray!10}0.892 &\cellcolor{gray!10}0.894 & \cellcolor{gray!10}0.962 \\ 
             \bottomrule
        \end{tabular}
    \end{adjustbox}
\end{table}

The baseline models demonstrate strong capability in reproducing the annotated GRS, achieving $\rho_c$ close to or greater than~0.8 on test sets for all tasks except circle cutting.
The comparatively poor performance on the circle cutting task is likely related to the low inter-rater agreement for this task (see Table~\ref{tab:agreement}), suggesting that the annotated scores are less reliable. Consequently, the validity of the skill ratings for circle cutting should be regarded as limited.

As expected, the ensemble generally outperforms individual models, yielding higher agreement with the annotated scores.  
Interestingly, models occasionally achieve better results on the test set than on the validation set.
This phenomenon can be explained by differences in the variance of the annotated GRS affecting the size of Lin's~$\rho_c$, which can be expected to increase with the variance of the annotated scores. 

Incorporating information about the task completion time provides an additional performance boost, particularly for the suture \& knot task. Yet, the completion time itself shows a strong negative correlation with the annotated GRS in several tasks, implying that faster execution indicates a higher total GRS. In particular, for suture \& knot, Spearman’s rank correlation coefficient~$\rho$ between video duration and GRS is -0.91; for peg transfer and circle cutting, $\rho$ equals -0.80 and -0.72, respectively. This observation aligns with expectations as these tasks are derived from the MISTELS curriculum, which only comprises tasks for which shorter completion times imply higher surgical skill \cite{derossis1998development}.
However, researchers should be aware that this strong connection may lead models to exploit video duration as a shortcut instead of learning meaningful video-based skill representations. Notably, balloon resection shows a much weaker correlation between duration and GRS ($\rho=\text{-}0.47$), indicating that completion time alone is not a reliable measure of skill across all tasks.

\subsubsection*{Video-level error recognition}

To further establish a baseline for automatic error recognition in surgical videos, we employ the same approach as illustrated in Fig.~\ref{fig:baseline_model}. Unlike the skill regression setup, the model here is trained as a binary classifier to compute whether an error occurred in a given recording, returning 1 for the presence of the error and 0 otherwise. 
In this setting, information about the video duration does not seem necessary.

For each task, we train baseline models to identify a single, well-represented error type: \texttt{object\_dropped\_within\_fov} for peg transfer, \texttt{gauze\_detached} for circle cutting, the combined error category \texttt{balloon\_opened OR balloon\_damaged}\footnote{
        This error appears in approximately 47\% of all balloon resection recordings.
    } 
for balloon resection, and \texttt{needle\_dropped} for suture \& knot.
To account for stochastic variability, training is repeated three times with different random seeds.
The \texttt{object\_dropped\_within\_fov} and \texttt{needle\_dropped} errors are typically visible in only a few clips within each video. To better capture these localized error events, we replace the global average pooling with global max pooling in the corresponding models.

For evaluation, we suggest to use two metrics: accuracy and balanced accuracy (the average of sensitivity and specificity), where the latter provides a fairer assessment when error distributions are imbalanced. Table~\ref{tab:automatic_error_recognition} presents mean and standard deviation of both metrics across the three runs, computed on both the validation and the test set.

\begin{table}[htb]
    \caption{Performance of the baseline models for automatic video-level error recognition on the LASANA validation and test sets. Reported values denote the mean and standard deviation over three independent runs.
    }
    \label{tab:automatic_error_recognition}
    \centering
    \begin{adjustbox}{width=\linewidth}
        \begin{tabular}{lllllllll}
            \toprule
            \multirow{2}{*}{Metric} & \multicolumn{2}{l}{\makecell[cl]{\texttt{object\_dropped}\\\texttt{\_within\_fov}}} & \multicolumn{2}{l}{\texttt{gauze\_detached}} & \multicolumn{2}{l}{\makecell[cl]{\texttt{balloon\_opened}\\\texttt{OR balloon\_damaged}}} & \multicolumn{2}{l}{\texttt{needle\_dropped}} \\
              & validation & test & validation & test & validation & test & validation & test \\ 
             \midrule
             Accuracy & $1.000_{\pm 0.000}$ & $0.833_{\pm 0.000}$ & $0.877_{\pm 0.021}$ & $0.880_{\pm 0.040}$ & $0.944_{\pm 0.019}$ & $0.922_{\pm 0.034}$ & $0.948_{\pm 0.048}$ & $0.820_{\pm 0.000}$ \\
             Balanced accuracy & $1.000_{\pm 0.000}$ & $0.815_{\pm 0.005}$ & $0.864_{\pm 0.019}$ & $0.857_{\pm 0.048}$ & $0.942_{\pm 0.017}$ & $0.922_{\pm 0.034}$ & $0.948_{\pm 0.047}$ & $0.820_{\pm 0.002}$ \\
             \bottomrule
        \end{tabular}
    \end{adjustbox}
\end{table}

The baseline models perform well in recognizing the annotated errors, achieving balanced accuracies above 80\% for all considered error types. In particular, the model for the balloon resection task demonstrates strong performance, reliably identifying whether the balloon was damaged or opened with an accuracy exceeding 90\%.

\subsubsection*{Implementation and training details}

Each baseline model is implemented in Python using PyTorch \cite{paszke2019pytorch} and trained on a single NVIDIA RTX A5000 GPU with 24\,GB VRAM. Video clips contain 16 consecutive frames with a temporal resolution of 4~fps, which are resized to $224 \times 224$ pixels. 
During inference, clips are extracted with a slight overlap between consecutive clips, using a fixed step size. 
During training, a randomized segment-based sampling strategy \cite{wang2016temporal} is employed, where the number of segments~$k$ is determined dynamically such that the selected clips collectively cover 75\% of the video. 
Data augmentation is implemented using the Albumentations library \cite{buslaev2020albumentations} to apply randomized transformations, including color and geometric modifications. All frames within a given clip undergo identical transformations to preserve temporal consistency.

For skill score regression, the 3D~CNN is initialized with Kinetics-pretrained weights \cite{carreira2017quo} and the full model is trained to regress the GRS. Training uses Huber loss \cite{huber1992robust} ($\delta = 0.5$) and the Adam optimizer \cite{kingma2014adam}. It is conducted for up to 300 epochs, with a batch size of one video or $k$ clips.
During training, the learning rate is kept constant\footnote{
To determine a good learning rate, we perform a grid search over $\lbrace{1 \cdot 10^{-5}, 3 \cdot 10^{-5},  5\cdot 10^{-5}, 1\cdot 10^{-4} \rbrace}$. The learning rate that achieves the best performance on the validation set, measured by Lin's $\rho_c$, is chosen. 
% For models using only video as input, we pick a \emph{lr} of $5\cdot 10^{-5}$ for suture \& knot and $3\cdot 10^{-5}$ for the remaining tasks. For models incorporating both video and task completion time, we pick a \emph{lr} of $3\cdot 10^{-5}$ for circle cutting and $5\cdot 10^{-5}$ for the remaining tasks.
} and the model checkpoint achieving the highest validation $\rho_c$ is kept for final evaluation on the test set. 
For long videos, where the number $k$ of extracted clips is large, gradients for updating the 3D~CNN are computed sequentially on smaller micro-batches \cite{liu2023etad}.

For error recognition, the training setup is similar. The only modifications include adopting a binary cross-entropy loss and selecting the model checkpoint that yields the highest validation accuracy instead of validation~$\rho_c$. % Following learning rate tuning (grid search) on the validation set, we set the \emph{lr} to $3 \cdot 10^{-5}$ for the \texttt{needle\_dropped} error and $5 \cdot 10^{-5}$ for the remaining error types. 

\section*{Usage Notes}

\subsection*{Data download}

All files are listed on the dataset webpage (\url{https://doi.org/10.25532/OPARA-1046}) and can be downloaded in the web browser of your choice. 
To download large zip files (up to 27\,GB) you can alternatively try the \texttt{wget}\footnote{\url{https://www.gnu.org/software/wget/}} utility:
\begin{verbatim}
cd <target_dir>
wget -c --tries=0 --timeout=30 -O <target_file> <url>
\end{verbatim}
Here, \texttt{<target\_dir>} refers to the folder that should store the file and \texttt{<target\_file>} is the name under which the file should be stored (e.g. \texttt{BalloonResection\_left.zip}). Without this option, the file will be stored under the default name \texttt{download}. The link to the file is denoted as \texttt{<url>} and can be obtained from the dataset webpage (e.g. in  Mozilla Firefox: right click > Copy Link). 
The option \texttt{-c} helps to download only the remainder of the file if the download is interrupted and resumed at a later point. 
 
\subsection*{Benchmarking}

When benchmarking a method for video-based skill assessment or error recognition, we suggest to adhere to the pre-defined data splits. The validation set should be used to refine the method during development, whereas the test set should only be used for the final evaluation. Ideally, results are reported for both the validation and the test sets.

The circle cutting task should not be used to evaluate approaches for automatic skill assessment.

\section*{Data Availability}

The LASANA dataset is available in the Open Access Repository and Archive for Research Data of Saxon Universities: \url{https://doi.org/10.25532/OPARA-1046}.

\section*{Code Availability}

\begin{itemize}
    \item Code for training and evaluating the baseline method for skill score regression and error recognition on the LASANA benchmark is provided at \url{https://gitlab.com/nct_tso_public/LASANA/lasana}.
    \item Code for analyzing the validity of the LASANA annotations, including the inter-rater agreement, and for creating illustrative plots is provided at \url{https://gitlab.com/nct_tso_public/LASANA/data_analysis}. 
    \item Code for stereo matching, using left and right video and the given camera calibration information, is provided at \url{https://gitlab.com/nct_tso_public/LASANA/dense_matching}.
\end{itemize}

\section*{Author Contributions}

\textbf{Isabel Funke:} Conceptualization, Investigation, Data Curation, Methodology, Software, Validation, Visualization, Writing--Original Draft
\textbf{Sebastian Bodenstedt:} Conceptualization, Software, Investigation
\textbf{Felix von Bechtolsheim:} Conceptualization, Project administration, Resources, Writing--Review \& Editing
\textbf{Florian Oehme:} Conceptualization, Project administration, Resources
\textbf{Michael Maruschke:} Investigation, Project administration
\textbf{Stefanie Herrlich:} Investigation, Project administration
\textbf{Jürgen Weitz:} Supervision, Resources
\textbf{Marius Distler:} Supervision, Resources
\textbf{Sören Torge Mees:} Conceptualization, Supervision, Resources
\textbf{Stefanie Speidel:} Conceptualization, Supervision, Resources, Funding acquisition, Writing--Review \& Editing.
\textbf{All authors} read and approved the final manuscript.

\section*{Competing Interests} 

The authors declare no competing interests.
 
\section*{Acknowledgments} 

The authors deeply thank the raters who provided annotations for the videos, all medical students and clinicians who contributed recordings to the dataset, all surgeons who were involved in teaching the course on minimally invasive surgery, and the research assistants who helped record the videos. Special thanks to Adamantini Hatzipanayioti who also assisted with video recording.

\section*{Funding} 

Funded by the German Research Foundation (DFG, Deutsche Forschungsgemeinschaft) as part of Germany’s Excellence Strategy – EXC 2050/1 – Project ID 390696704 – Cluster of Excellence ``Centre for Tactile Internet with Human-in-the-Loop'' (CeTI) of Technische Universität Dresden.



\section*{Supplementary material}

# Annotation guidelines

## General instructions

- Each video is rated on four aspects: depth perception, efficiency, bimanual dexterity, and tissue handling.

- Each aspect is independently assessed on a 5-point Likert scale, ranging from 1 (very bad) to 5 (very good).

- For each aspect, task-specific anchor descriptors are provided to define the rating scale.

## Task-specific instructions

### Peg transfer

***Task description***

Six triangular-shaped rubber objects must be transferred from the left to the right side of a pegboard, using two laparoscopic graspers. This means that each object needs to be picked up with the grasper in the left hand, transferred mid-air to the grasper in the right hand, and then placed on an arbitrary peg on the right side of the pegboard. Afterward, the objects must be transferred from the right side back to the left side, picking each object up with the right hand and, after transfer, placing it on a peg with the left hand.

***Error description***

- A triangular object is dropped within the field of view, and it can be retrieved.

- A triangular object is dropped outside the field of view, or it cannot be retrieved.

***Criteria for exclusion from the dataset***

- Task duration > 6 min.

***Rating scale with anchor descriptors***

*Depth perception*

| 1 | 2 | 3 | 4 | 5 |
|---|---|---|---|---|
| • Objects are approached very slowly and missed multiple times<br><br>• Transferring objects from one grasper to the other often requires several attempts<br><br>• Placing the objects takes twice as long as picking them up | | • Objects are approached at moderate speed and are not missed<br><br>• Transferring objects from one grasper to the other works at a moderate pace<br><br>• Placing the objects works without any issues | | • Objects are approached purposefully and quickly, and they are not missed<br><br>• Transferring objects from one grasper to the other is efficient and fast<br><br>• Objects are not dropped or placed next to a peg |

*Efficiency*

| 1 | 2 | 3 | 4 | 5 |
|---|---|---|---|---|

**1**
- Picking up the first object is unnecessarily delayed
- For the transfer, the second grasper only moves toward the object after the first grasper has picked it up
- Placing an object is hesitant, and the object needs to be aligned in the process

**3**
- Picking up the first object is slow but deliberate
- Objects are transferred directly from one grasper to the other, without the need to realign them
- Objects are placed in a directed manner, with slight alignment movements required

**5**
- The first object is picked up immediately and without issues
- Picking up and placing the objects appears rehearsed, with no interruptions
- Placing the objects does not require alignment movements
- No movement is too long; every movement has a clear goal

*Bimanual dexterity*

| 1 | 2 | 3 | 4 | 5 |
|---|---|---|---|---|

**1**
- An object is picked up by one grasper while the other grasper stays at the opposite side of the pegboard
- Only one grasper moves to reach the position for transfer
- During the transfer, the graspers do not interact and do not move toward the target peg

**3**
- An object is picked up by one grasper while the other grasper moves towards the position for transfer
- During the transfer, the graspers interact but remain in place and do not move toward the target peg
- Objects are placed purposefully, while the grasper used for pick-up remains in place

**5**
- An object is picked up by one grasper while the other one already waits at the position for transfer
- During the transfer, the graspers interact and move toward the target peg
- Objects are placed purposefully, while the grasper used for pick-up is already moving back to the opposite side of the pegboard

*Tissue handling*

| 1 | 2 | 3 | 4 | 5 |
|---|---|---|---|---|

**1**
- Objects are strongly squeezed
- Objects are not grasped properly and securely, and without making use of the holes in the objects

**3**
- Objects are properly grasped and not squeezed
- Some pressure is applied to the objects, for example when placing them on the pegs

**5**
- Objects are properly grasped and not squeezed
- Throughout the task, no pressure is applied to the objects, neither when picking them up nor when placing them on the pegs

## Circle cutting

### *Task description*

A two-ply piece of gauze is marked with a circle and secured with four metal clips. Using a grasper and one pair of laparoscopic scissors, the circle must be cut out from the gauze piece, cutting exactly along the marked line. Here, it is irrelevant how much or little of the bottom layer of the gauze is cut. Instruments may be switched between hands at any time.

### *Error description*

- Cuts are made inside or outside the marked circle.
- The gauze comes out of one (or more) of the metal clips.

### *Criteria for exclusion from the dataset*

- Task duration > 10 min.
- Gauze is of unsuitable, very coarse material.

### *Rating scale with anchor descriptors*

*Depth perception*

| 1 | 2 | 3 | 4 | 5 |
|---|---|---|---|---|
| • At the beginning, the gauze must be grasped multiple times to find the starting point<br>• The scissors approach slowly and pass the target at least once<br>• Cutting motions are repeated because the scissors cut the air rather than the gauze | | • The grasper reaches the gauze slowly but steadily<br>• In more than half of the cases, the scissors actually cut the gauze<br>• The repositioning of the grasper is slow but precise | | • At the beginning, scissors and grasper purposefully reach the gauze immediately<br>• The gauze is cut with every closure of the scissors<br>• Repositioning of the grasper works quickly and without the need to readjust |

*Efficiency*

| 1 | 2 | 3 | 4 | 5 |
|---|---|---|---|---|
| • At the beginning, the gauze must be grasped multiple times to find the starting point<br>• The scissors are not properly aligned and need to be re-aligned multiple times<br>• Cuts are made repeatedly at the same spot | | • The gauze is reached slowly but steadily<br>• The starting point of the cut is maintained; it is avoided to create multiple new starting points<br>• Progress is made with more than half of the cuts<br>• The grasper is used in a supportive way | | • The gauze is reached quickly with both instruments<br>• The starting point of the cut is defined with the first cut and is maintained afterward<br>• Progress is made with every cut<br>• Repositioning of the instruments is deliberate and intentional |

## Bimanual dexterity

| 1 | 2 | 3 | 4 | 5 |
|---|---|---|---|---|
| • The instruments move only one after the other<br><br>• The grasper does not follow up on time to assist the scissors<br><br>• The grasper is barely used to support the cutting process and/or interferes with the scissors<br><br>• The instruments are switched between hands multiple times | | • Scissors and grasper assist each other and stay close to one another<br><br>• The grasper does not interfere with the scissors<br><br>• The grasper follows up to enable the scissors to work effectively, but moves with a delay<br><br>• The instruments are switched between hands only when it is reasonable | | • Both instruments move simultaneously, quickly, and purposefully<br><br>• The grasper follows up as soon as the scissors can no longer cut the gauze easily<br><br>• The grasper pulls the gauze in the correct direction even from difficult positions<br><br>• The instruments are switched between hands at most once |

## Tissue handling

| 1 | 2 | 3 | 4 | 5 |
|---|---|---|---|---|
| • The first grip is rough and creates tension on the gauze<br><br>• The gauze is torn; the grasper tears fibers<br><br>• Strong tension on the gauze causes it to be pulled out of one of the metal clips<br><br>• The circle is cut out imprecisely | | • The gauze does not tear; the fibers remain intact<br><br>• Some tension is applied to the gauze while gripping<br><br>• The circle is cut out with some minor deviations | | • No unnecessary tension is applied to the gauze; fibers are not torn<br><br>• The circle is cut out precisely<br><br>• The depth of the cuts is consistent |

## Balloon resection

### *Task description*

A standard latex balloon, filled with water, is inserted into another balloon that is marked with a line in reverse T shape. Both balloons are closed with a common knot and mounted with a metal clip at the knotted end. The inner balloon must be removed unbroken from the outer balloon by accurately cutting only the outer layer along the marked line. Instruments may be switched between hands at any time.

### *Error description*

- Cuts deviate from the marked line.

- The marked line is not completely cut.

- The inner balloon is opened, which means that more than half of the water leaks out. If the inner balloon is opened with the first cut already, the task should be marked as *failed*.

- The inner balloon is damaged, but no more than half of the water leaks out.

- The inner balloon is perforated, which means that only single drops of water leak out or water only leaks out when pressure is applied to the balloon.

### *Criteria for exclusion from the dataset*

- Task duration > 10 min.

- Light of endoscope is set to low.

### *Rating scale with anchor descriptors*

<p align="center"><em>Depth perception</em></p>

| 1 | 2 | 3 | 4 | 5 |
|---|---|---|---|---|
| • Multiple attempts are required to grasp the balloons<br><br>• The scissors approach slowly; the first cut is made only after several attempts<br><br>• Very slowly, the scissors are re-aligned with the cutting plane before each cut | | • The balloons are grasped purposefully but slowly<br><br>• The scissors approach slowly but steadily<br><br>• After some initial probing, the first cut is successful and is performed with care | | • The balloons are grasped quickly; the outer layer is immediately placed under tension to detach the inner balloon<br><br>• The scissors approach immediately, making the first cut quickly and purposefully<br><br>• The scissors remain in the correct cutting plane<br><br>• Repositioning of both instruments is quick and precise |

## Efficiency

| 1 | 2 | 3 | 4 | 5 |
|---|---|---|---|---|

- Multiple attempts are required to grasp the balloons at the beginning
- The opening in the outer layer is lost during cutting, meaning that the cutting plane must be reestablished
- The grasper does not support the scissors
- The scissors are repositioned very often

- Fewer than two attempts are required to grasp the balloons at the beginning
- The grasper assists before each cut
- Cutting the outer layer appears smooth
- The overall procedure appears fluid, with no noticeable pauses

- The balloons are grasped only once at the beginning; the outer layer is opened with the first cut
- The grasper holds the cutting plane in place before each cut
- The scissors cut continuously without having to be repositioned

## Bimanual dexterity

| 1 | 2 | 3 | 4 | 5 |
|---|---|---|---|---|

- The second hand does not help to stabilize the balloons for the initial grasp
- The grasper does not support the cutting process and/or interferes with the scissors

- The second hand slowly helps to stabilize the balloons for the initial grasp
- The grasper supports most of the cutting process, staying in contact with the outer layer at least half of the time
- The grasper rarely interferes with the scissors

- The second hand quickly helps to stabilize the balloons for the initial grasp
- The grasper continuously supports the cutting process, staying in contact with the outer layer most of the time
- The grasper does not interfere with the scissors

## Tissue handling

| 1 | 2 | 3 | 4 | 5 |
|---|---|---|---|---|

- The initial grasp is rough, without careful exposure of the outer layer
- The outer layer is torn open several times by excessive tension
- The outer layer is cut roughly and without respecting the marked line
- The inner balloon is opened, which means that more than half of the water leaks out

- The initial grasp is gentle
- The first cut is made after tension is applied to the outer layer to detach the inner balloon
- The outer layer is not torn
- Most of the cuts do not deviate from the marked line

- The outer layer is safely and visibly separated from the inner layer using fine, careful tension by the grasper
- The first cut is very delicate and minimally opens the outer layer
- The grasper applies well-controlled tension to the outer layer
- The cuts do not deviate from the marked line
- The inner balloon is not damaged

## Laparoscopic suture and knot

### Task description

A Penrose drain with a slit and a marked dot on each side of the slit is secured on a Velcro strip. The slit in the Penrose drain must be closed by placing a suture of length 15 cm accurately through the two marked dots and tying a knot. The knot must consist of one double throw, followed by two single throws. The laparoscopic instruments provided for this task are one needle driver to hold the needle end of the suture and one grasper.

*Note*: The maximum time to complete this task is 10 min, which means that the video recording stops after 10 min.

### Error description

- The needle is dropped.

- The suture deviates from any of the two marked dots on the Penrose drain.

- Within 10 min, fewer than three throws are successfully completed. In this case, the task should be marked as *failed*.

- The slit in the Penrose drain is not closed properly.

- The knot slips or comes apart.

- The Penrose drain is separated from the Velcro strip.

### Criteria for exclusion from the dataset

- Within 10 min, no throw can be completed.

### Rating scale with anchor descriptors

*Depth perception*

| 1 | 2 | 3 | 4 | 5 |
|---|---|---|---|---|

- Grasping needle or thread takes multiple attempts
- The needle approaches the Penrose drain very slowly and misses the marked dots
- Creating loops of thread requires multiple attempts

- Needle and thread are grasped slowly but reliably
- The needle approaches the Penrose drain slowly, but hits the marked dots
- Creating loops of thread is slow but requires only few attempts

- Needle and thread are grasped quickly and reliably
- The needle is pushed directly through the marked dots, with no corrective movements
- Loops of thread are created flawlessly and purposefully
- Loops are tightened precisely, without readjustments

*Efficiency*

| 1 | 2 | 3 | 4 | 5 |
|---|---|---|---|---|

- The needle must be regrasped or reoriented before stitching
- After stitching, the needle is grasped incorrectly and must be reoriented
- The needle is passed between instruments very frequently

- The needle only requires minor adjustments before stitching
- Stitching is done slowly but steadily

- The needle does not need to be reoriented before stitching
- Stitching into and out of the Penrose drain can be done in one movement without correction
- The grasper takes over the needle immediately after stitching, and the first throw can be started instantly

## Bimanual dexterity

| 1 | 2 | 3 | 4 | 5 |
|---|---|---|---|---|

**1**
- Only one instrument is used to orient the needle
- The grasper does not grab the Penrose drain for better visualization of the marked dots
- The needle driver both pushes the needle through and pulls the needle out of the Penrose drain; the grasper is not used at all
- The instruments interfere with each other

**3**
- One instrument primarily performs the orientation of the needle; the second one assists only occasionally
- During stitching, the grasper takes over only after the motion of the needle driver is finished
- When throwing the thread, only one instrument moves while the other remains static

**5**
- Both instruments assist in orienting the needle
- The grasper grabs the Penrose drain to visualize the marked dots
- During stitching, the needle driver rotates the needle to a position where the grasper can take over directly
- When throwing the thread, both instruments move together
- Both instruments interact to pass the needle between hands

## Tissue handling

| 1 | 2 | 3 | 4 | 5 |
|---|---|---|---|---|

**1**
- When orienting the needle, the needle is bent and/or the thread is torn
- The needle is pushed straight through the tissue instead of being rotated in a semicircular motion
- The Penrose drain is damaged when the thread is pulled through

**3**
- The needle is oriented with little force and little tension on the thread
- The Penrose drain is briefly under tension, with little tearing of the material
- Loops are tightened forcefully but with low tension

**5**
- When orienting the needle, the needle is not bent and the thread is not torn
- The Penrose drain experiences minimal tension during stitching and no tearing when the thread is pulled through
- Loops are tightened without tearing or excessive tension



\end{document}